\documentclass[10pt,twocolumn,letterpaper]{article}

\usepackage{cvpr}

\usepackage{times}
\usepackage{epsfig}
\usepackage{graphicx}
\usepackage{amsmath}
\usepackage{amssymb}
\usepackage{mathtools}
\usepackage{subcaption}
\usepackage{multirow}
\usepackage[export]{adjustbox}

\usepackage[colorinlistoftodos]{todonotes}

\usepackage[nolist]{acronym}

\usepackage[breaklinks=true,bookmarks=false]{hyperref}

\let\tanh\relax

\DeclareMathOperator\tanh{\eta}

\DeclareMathOperator\softmax{softmax}

\DeclareMathOperator*{\argmax}{arg\,max}
\newcommand{\io}[1]{{\bf #1}}
\def\cnv{*}

\graphicspath{{figs/}}

\newacro{lstm}[LSTM]{Long Short-Term Memory}
\newacro{lsta}[LSTA]{Long Short-Term Attention}
\newacro{clstm}[ConvLSTM]{Convolutional Long Short-Term Memory}

\cvprfinalcopy 

\def\cvprPaperID{5009} 

\definecolor{mygc}{RGB}{0,179,80}
\definecolor{myrc}{RGB}{255,0,0}
\newcommand{\myg}{\color{mygc}}
\newcommand{\myr}{\color{myrc}}

\begin{document}

\title{\acs{lsta}: Long Short-Term Attention
for Egocentric Action Recognition}


\author{Swathikiran Sudhakaran$^{1,2}$, Sergio Escalera$^{3,4}$, Oswald Lanz$^{1}$\\ 
	$^{1}$Fondazione Bruno Kessler, Trento, Italy\\
	$^{2}$University of Trento, Trento, Italy\\
	$^{3}$Computer Vision Center, Barcelona, Spain\\
	$^{4}$Universitat de Barcelona, Barcelona, Spain\\
	{\tt\small \{sudhakaran,lanz\}@fbk.eu, \tt\small sergio@maia.ub.es}
}

\maketitle

\begin{abstract}
Egocentric activity recognition is one of the most challenging tasks in video analysis. It requires a fine-grained discrimination of small objects and their manipulation. While some methods base on strong supervision and attention mechanisms, they are either annotation consuming or do not take spatio-temporal patterns into account. In this paper we propose LSTA as a mechanism to focus on features from relevant spatial parts while attention is being tracked smoothly across the video sequence. We demonstrate the effectiveness of LSTA on egocentric activity recognition with an end-to-end trainable two-stream architecture, achieving state-of-the-art performance on four standard benchmarks.

\end{abstract}

\vspace{-0.5cm}
\section{Introduction}

Recognizing human actions from videos is a widely studied  problem in computer vision. Most research is devoted to the analysis of videos captured from distant, third-person views. Egocentric (first-person) video analysis is an important and relatively less explored branch with potential applications in robotics, indexing and retrieval, human-computer interaction, or human assistance, just to mention a few. Recent advances in deep learning highly benefited problems such as image classification \cite{he16residual, xie2017aggregated} and object detection \cite{liu2016ssd, he2017mask}. However, the performance of deep learning action recognition from videos is still not comparable to the advances made in object recognition from still images \cite{he16residual}. One of the main difficulties in action recognition is the huge variations present in the data caused by the highly articulated nature of the human body. Human kinesics, being highly flexible in nature, results in high intra-subject and low inter-subject variabilities. This is further challenged by the variations introduced by the unconstrained nature of the environment where the video is captured. Since videos are composed of image frames, this introduces an additional dimension to the data, making it more difficult to define a model that properly focuses on the regions of interest that better discriminate particular action classes. In order to mitigate these problems, one approach could be the design of a large scale dataset with fine-grain annotations covering the space of spatio-temporal variabilities defined by the problem domain, which would be unfeasible in practice.

Here, we consider the problem of identifying fine-grained egocentric activities from trimmed videos. This is a comparatively difficult task considered to action recognition since the activity class depends on the action and the object on to which the action is applied to. This requires the development of a method that can simultaneously recognize the action as well as the object. In addition, the presence of strong ego-motion caused by the sharp movements of the camera wearer introduces noise to the video that complicates the encoding of motion in the video frame. While incorporating object detection can help the task of egocentric action recognition, still this would require fine-grain frame level annotations, becoming costly and impractical in a large scale setup. 
	
Attention in deep learning was recently proposed to guide networks to focus on regions of interest relevant for a particular recognition task. This prunes the network search space and avoids computing features from irrelevant image regions, resulting in a better generalization. Existing works explore both bottom-up \cite{attention_eccv18} and top-down attention mechanisms \cite{sudhakaran2018attention}. Bottom-up attention relies on the salient features of the data and is trained to identify such visual patterns that distinguish one class from another. Top-down attention applies prior knowledge about the data for developing attention, e.g. the presence of certain objects which can be obtained from a network trained for a different task. Recently, attention mechanisms have been successfully applied to egocentric action recognition \cite{li2018eye, sudhakaran2018attention}, surpassing the performance of non-attentive alternatives. Still, very few attempts have been done to track attention into spatio-temporal egocentric action recognition data. As a result, current models may lose a proper smooth tracking of attention regions in egocentric action videos. Furthermore, most current models base on separate pre-training with strong supervision, requiring complex annotation operations.  

To address these limitations, in this work we investigate on the more general question of \emph{how a video CNN-RNN can learn to focus on the regions of interest to better discriminate} the action classes. We analyze the shortcomings of \acsp{lstm} in this context and derive \ac{lsta}, a new recurrent neural unit that augments \acs{lstm} with built-in spatial attention and a revised output gating. The first enables \ac{lsta} to attend the feature regions of interest while the second constraints it to expose a distilled view of internal memory. 
Our study confirms that it is effective to improve the output gating of recurrent unit since it does not only affect prediction overall but controls the recurrence, being responsible for a smooth and focused tracking of the latent memory state across the sequence. Our main contributions can be summarized as follows:
\begin{itemize}
    \item We present \acf{lsta}, a new recurrent unit that addresses shortcomings of \acs{lstm} when the discriminative information in the input sequence can be spatially localized;\vspace{-0.2cm}
    \item We deploy \ac{lsta} into a two stream architecture with cross-modal fusion, a novel control of the bias parameter of one modality by using the other\footnote{Code is available at \href{https://github.com/swathikirans/LSTA}{https://github.com/swathikirans/LSTA}};\vspace{-0.2cm}
    \item We report an ablation analysis of the model and evaluate it on egocentric activity recognition, providing state-of-the-art results in four public datasets.
\end{itemize}

\section{Related Work}
\label{sec:related}
We discuss the most relevant deep learning methods for addressing egocentric vision problems in this section.

\subsection{First Person Action Recognition}

The works of \cite{ma2016deeper, singh2016first, zhou2016cascaded} train specialized CNN for hand segmentation and object localization related to the activities to be recognized. These methods base on specialized pre-training for hand segmentation and object detection networks, requiring high amounts of annotated data for that purpose. Additionally, they just base on single RGB images for encoding appearance without considering temporal information. In \cite{ryoo2015pooled, zaki2017modeling} features are extracted from a series of frames to perform temporal pooling with different operations, including max pooling, sum pooling, or histogram of gradients. Then, a temporal pyramid structure allows the encoding of both long term and short term characteristics. However, all these methods do not take into consideration the temporal order of the frames. Techniques that use a recurrent neural network such as \acf{lstm} \cite{cao2017egocentric, verma2018making} and \acf{clstm} \cite{sudhakaran2017convolutional, sudhakaran2018attention} are proposed to encode the temporal order of features extracted from a sequence of frames. Sigurdsson \etal \cite{sigurdsson2018actor} proposes a triplet network to develop a joint representation of paired third person and first person videos. Their method can be used for transferring knowledge from third person domain to first person domain thereby partially solving the problem of lack of large first person datasets. Tang \etal \cite{tang2017action, tang2018multi} add an additional stream that accepts depth maps to the two stream network
enabling it
to encode 3D information present in the scene. Li \etal \cite{li2018eye} propose a deep neural network to jointly predict the gaze and action from first person videos, which requires gaze information during training. 

Majority of the state-of-the-art techniques rely on additional annotations such as hand segmentation, object bounding box or gaze information. This allows the network to concentrate on the relevant regions in the frame and helps in distinguishing each activity from one another better. However, manually annotating all the frames of a video with these information is impractical. For this reason, development of techniques that can identify the relevant regions of a frame without using additional annotations is crucial.

\subsection{Attention}

Attention mechanism was proposed for focusing attention on features that are relevant for the task to be recognized. This includes \cite{sudhakaran2018attention, li2018eye, shen2018egocentric} for first person action recognition, \cite{anderson2018bottom, ma2017attend, wang2018bidirectional} for image and video captioning and \cite{nguyen2018improved, anderson2018bottom, liang2018focal} for visual question answering. The works of \cite{sharma2015action, girdhar2017attentional, sudhakaran2018top, sudhakaran2018attention, attention_eccv18, li2018eye} use an attention mechanism for weighting spatial regions that are representative for a particular task. Sharma \etal \cite{sharma2015action} and Zhang \etal \cite{attention_eccv18} generate attention masks implicitly by training the network with  video labels. 
Authors of \cite{girdhar2017attentional, sudhakaran2018top, sudhakaran2018attention} use top-down attention generated from the prior information encoded in a CNN pre-trained for object recognition while \cite{li2018eye} uses gaze information for generating attention.  The work of \cite{piergiovanni2017learning, shen2018egocentric} uses attention for weighting relevant frames, thereby adding temporal attention. This is based on the idea that not all frames present in a video are equally important for understanding the action being carried out. In \cite{piergiovanni2017learning} a series of temporal attention filters is learnt that weight frame level features depending on their relevance for identifying actions. \cite{shen2018egocentric} uses change in gaze for generating the temporal attention. \cite{li2018videolstm, du2018recurrent} apply attention on both spatial and temporal dimensions to select relevant frames and the regions present in them.

Most 
existing techniques for generating spatial attention in videos consider each frame independently. 
Since video frame sequences have an absolute temporal consistency, per frame processing results in the loss of valuable information.

\subsection{Relation to state-of-the-art alternatives}

The proposed LSTA method generates the spatial attention map in a top-down fashion utilizing prior information encoded in a CNN pre-trained for object recognition and another pre-trained for action recognition.   \cite{sudhakaran2018attention} proposes a similar top-down attention mechanism. However, they generate the attention map independently in each frame whereas in the proposed approach, the attention map is generated in a sequential manner. This is achieved by propagating the attention map generated from past frames across time by maintaining an internal state for attention. Our method uses attention on the motion stream followed by a cross-modal fusion of the appearance and motion streams, thereby enabling both streams to interact earlier in the layers to facilitate flow of information between them. \cite{attention_eccv18} proposes an attention mechanism that takes in to consideration the inputs from past frames. Their method is based on bottom-up attention and generates a single weight matrix which is trained with the video level label. However, the proposed method generates attention, based on the input, from a pool of attention maps which are learned using video level label alone.



\section{Analysis of \ac{lstm}}
\label{sec:lstm}
\ac{lstm} is the widely adopted neuron design for processing and/or predicting sequences. A latent memory state $\io{c_t}$ is tracked across a sequence with a forget-update mechanism
\begin{equation}
  \io{c_t} = f \odot \io{c_{t-1}} + i \odot c\label{eq:lstm.0}
\end{equation}
where $(f,i)$ have a gating function on the previous state $\io{c_{t-1}}$ and an innovation term $c$. $(f,i,c)$ are parametric functions of input $\io{x_t}$ and a gated non-linear view of previous memory state $\io{o_{t-1}} \odot \tanh(\io{c_{t-1}})$
\begin{equation}
(i,f,\io{o_t},c) = (\sigma,\sigma,\sigma,\tanh)(W[\io{x_t}, \io{o_{t-1}} \odot \tanh(\io{c_{t-1}})])\label{eq:lstm.1} 
\end{equation}
The latter, referred to as hidden state $\io{h_t}=\io{o_t} \odot \tanh(\io{c_t})$, is often exposed to realize a sequence prediction. For sequence classification instead, the final memory state can be used as a fixed-length descriptor of the input sequence.

Two features of \ac{lstm} design explain its success. First, the memory update (Eq.~\ref{eq:lstm.0}) is flexibly controlled by $(f,i)$: a state can, in a single iteration, be erased $(0,0)$, reset $(0,1)$, left unchanged $(1,0)$, or progressively memorize new input. $(1,1)$ resembles residual learning~\cite{he16residual}, a key design pattern in very deep networks - depth here translates to sequence length. Indeed, \acp{lstm} has strong gradient flow and learn long-term dependencies~\cite{Hochreiter:1997:LSM:1246443.1246450}. Second, the gating functions (Eq.~\ref{eq:lstm.1}) are learnable neurons and their interaction in memory updating is transparent (Eq.~\ref{eq:lstm.0}). When applied to video classification, a few limitations are to be discussed:\\
{1. \em Memory}. Standard \acp{lstm} use fully connected neuron gates and consequently, the memory state is unstructured. This may be desired \eg for image captioning where one modality (vision) has to be translated into another (language). For video classification it might be advantageous to preserve the spatial layout of images and their convolutional features by propagating a memory tensor instead. Conv\ac{lstm}~\cite{shi15convlstm} addresses this shortcoming through convolutional gates in the \ac{lstm}.\\
{2. \em Attention}. The discriminative information is often confined locally in the video frame. Thus, not all convolutional features are equally important for recognition. In \acp{lstm} the filtering of irrelevant features (and memory) is deferred to the gating neurons, that is, to a linear transformation (or convolution) and a non-linearity. Attention neurons were introduced to suppress activations from irrelevant features ahead of gating. We augment \ac{lstm} with built-in attention that directly interacts with the memory tracking in Sec.~\ref{sec:attention}.\\
{3. \em Output gating}. Output gating not only impacts sequence prediction but it critically affects memory tracking too, \emph{cf.} Eq~\ref{eq:lstm.1}. We replace the output gating neuron of \ac{lstm} with a high-capacity neuron whose design is inspired by that of attention. There is indeed a relation among them, we make this explicit in Sec.~\ref{sec:pooling}.\\
{4. \em External bias control}. The neurons in Eq.~\ref{eq:lstm.1} have a bias term that is learnt from data during training, and it is fixed at prediction time in standard \ac{lstm}. We leverage on adapting the biases based on the input video for each prediction. State-of-the-art video recognition is realized with two-stream architectures, 
we use flow stream to control appearance biases in Sec.~\ref{sec:twostream.cross}.

\section{Long Short-Term Attention}
\label{sec:lsta}

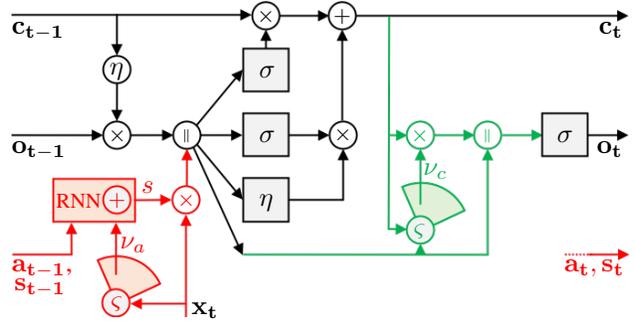
\begin{figure}[!t]
\centering\includegraphics[width=\linewidth]{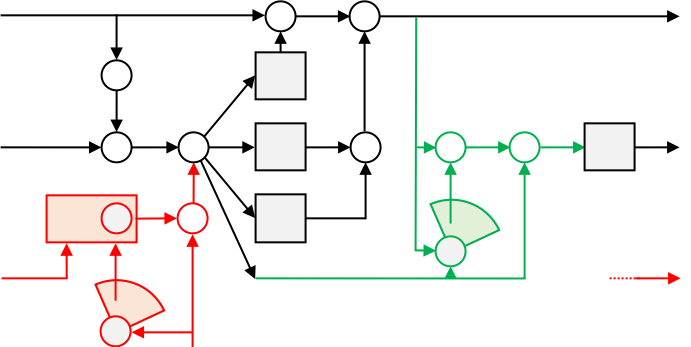}
\put(-237,107){$\io{c_{t-1}}$}
\put(-15,107){$\io{c_t}$}
\put(-237,62){$\io{o_{t-1}}$}
\put(-15.7,62){$\io{o_t}$}
\put(-28,17){\myr$\io{a_t},\io{s_t}$}
\put(-169,1){$\io{x_t}$} 
\put(-237,17){\myr$\io{a_{t-1}}$,}
\put(-236,10){\myr$\footnotesize\io{s_{t-1}}$}
\put(-200,91.5){$\eta$}
\put(-200.45,66.47){\footnotesize$\times$}
\put(-144.2,111.6){\footnotesize$\times$}
\put(-115.3,111.6){\footnotesize$+$}
\put(-115.1,66.6){\footnotesize$\times$}
\put(-143.5,90.5){$\sigma$}
\put(-143.5,66){$\sigma$}
\put(-143.5,42){$\eta$}
\put(-86,66.3){\myg\footnotesize$\times$}
\put(-30.8,66){$\sigma$}
\put(-200.5,41.9){\myr\footnotesize$+$}
\put(-174.3,41.95){\myr\footnotesize$\times$}
\put(-220.2,41.2){\myr\footnotesize RNN}
\put(-188,47){\myr$s$}
\put(-196,25.8){\myr$\nu_a$}
\put(-199.6,3.25){\myr$\varsigma$}
\put(-81.3,53.5){\myg$\nu_c$}
\put(-85,30.5){\myg$\varsigma$}
\put(-172.8,67){\tiny$\Vert$}
\put(-59,67){\myg\tiny$\Vert$}
\caption{\ac{lsta} extends \ac{lstm} with two novel components: recurrent attention and output pooling. The first (red part) tracks a weight map $s$ to focus on relevant features, while the second (green part) introduces a high-capacity output gate. 
At the core of both is a pooling operation $\varsigma$, that selects one out of a pool of specialized mappings to realize smooth attention tracking and flexible output gating. Circles indicate point-wise or concat operations, square blocks are linear/convolutional parametric nodes with non-linearities indicated by their symbols. Recurrent variables in bold.
}
\label{fig:lsta}\vspace{-0.5cm}
\end{figure}

We present a schematic view of \acs{lsta} in Fig.~\ref{fig:lsta}. \acs{lsta} extends \acs{lstm}~\cite{gers2000learning} with two newly designed components. The core operation is a pooling~$\varsigma$, that selects one out of a pool of specialized mappings to realize attention tracking (red part) and output gating (green part). The pooling $\varsigma$ on features $\io{x_t}$ returns a 
map $\nu_a$ that is fed
through a conventional RNN cell with memory $\io{a_t}$ and output gate $\io{s_t}$. Its output state $\io{s_t}\odot\tanh(\io{a_t})$ is added to the input $\nu_a$ and softmax calibrated to obtain an attention map $s$. The map $s$ is then applied to $\io{x_t}$, that is, $s\odot\io{x_t}$ is the attention filtered feature for updating memory state $\io{c_t}$ using conventional LSTM recurrence (black part). Our redesigned output gating  uses a filtered view of the updated memory state, $\nu_c\odot\io{c_t}$, instead of $\io{x_t}$. To obtain $\nu_c$ through pooling we use $s\odot\io{x_t}$ to control the bias of operator $\varsigma$, hereby coupling attention tracking with output gating. This model is instantiated for action recognition from egocentric video in its convolutional version as\\[-0.6cm]
\begin{eqnarray}
  \nu_a &\!\!\!\!=\!\!\!\!& \varsigma(\io{x_t}, w_a)\label{eq:lsta.0}\\
  (i_a,f_a,\io{s_t},a) &\!\!\!\!=\!\!\!\!& (\sigma,\sigma,\sigma,\tanh)(W_a \cnv [\nu_a, \io{s_{t-1}} \odot \tanh(\io{a_{t-1}})])\label{eq:lsta.1}\\
  \io{a_t} &\!\!\!\!=\!\!\!\!& f_a \odot \io{a_{t-1}} + i_a \odot a\label{eq:lsta.2}\\
  s &\!\!\!\!=\!\!\!\!& \softmax(\nu_a + \io{s_t} \odot \tanh(\io{a_t}))\label{eq:lsta.3}\\
  (i_c,f_c,c) &\!\!\!\!=\!\!\!\!& (\sigma,\sigma,\tanh)(W_c \cnv [s \odot \io{x_t}, \io{o_{t-1}} \odot \tanh(\io{c_{t-1}})])\label{eq:lsta.4}\\
  \io{c_t} &\!\!\!\!=\!\!\!\!& f_c \odot \io{c_{t-1}} + i_c \odot c\label{eq:lsta.5}\\
  \nu_c &\!\!\!\!=\!\!\!\!& \varsigma(\io{c_t}, w_c + w_o \epsilon(s \odot \io{x_t}))\label{eq:lsta.6}\\
  \io{o_t} &\!\!\!\!=\!\!\!\!& \sigma(W_o * [\nu_c\odot\io{c_t}, \io{o_{t-1}} \odot \tanh(\io{c_{t-1}})])\label{eq:lsta.7}
\end{eqnarray}

Eqs.~\ref{eq:lsta.0}-\ref{eq:lsta.3} implement our recurrent attention as detailed in Sec.~\ref{sec:attention}, Eqs.~\ref{eq:lsta.6}-\ref{eq:lsta.7} is our coupled output gating of Sec.~\ref{sec:pooling}. Bold symbols represent the recurrent variables: $(\io{a_t},\io{s_t})$ of shape $N \times 1$, $(\io{c_t},\io{o_t})$ of shape $N \times K$. Trainable parameters are: $(W_a,W_c)$ are both $K$ convolution kernels, $(w_a,w_c)$ have shape $K \times C$, $w_o$ has shape $C \times C$. $N,K,C$ are introduced below. $\sigma,\tanh$ are sigmoid and tanh activation functions, $*$ is convolution, $\odot$ is point-wise multiplication. $\varsigma,\epsilon$ are from the pooling model presented next. 

\subsection{Attention Pooling}
\label{sec:attention}

Given a matrix view $\io{x}_{ik}$ of convolutional feature tensor $\io{x}$ where $i$ indexes one of $N$ spatial locations and $k$ indexes one of $K$ feature planes, we aim at suppressing those activations $\io{x}_i$ that are uncorrelated with the recognition task. That is, we seek a $\varsigma(\io{x},w)$ of shape $1 \times N$ such that parameters $w$ can be tuned in a way that $\varsigma(\io{x},w) \odot \io{x}$ are the discriminative features for recognition. For egocentric activity recognition these can be from objects, hands, or implicit patterns representing object-hand interactions during manipulation.

Our design of $\varsigma(\io{x},w)$ is grounded on the assumption that there is a limited number of pattern categories that are relevant for an activity recognition task. Each category itself can, however, instantiate patterns with high variability during and across executions. We therefore want $\varsigma$ to select from a pool of category-specific mappings, based on the current input $\io{x}$. We want both the selector and the pool of mappings be learnable and self-consistent, and realized with fewer tunable parameters.

A selector with parameters $w$ maps an image features $\io{x}$ into a category-score space $\mathcal{C}$ from which the category $c^* \in \mathcal{C}$ obtaining the highest score is returned. Our selector is of the form $c^* = \argmax_c\pi(\epsilon(\io{x}),\theta_c)$ where $\epsilon$ is a reduction and $\theta_c \in w$ are the parameters for scoring $\io{x}$ against category $c$. If $\pi$ is chosen to be equivariant to reduction $\epsilon$ then $\pi(\epsilon(\io{x}), \theta_c) = \epsilon(\pi(\io{x}, \theta_c))$ and we can use $\{\epsilon^{\perp}(\pi(\cdot,\theta_c)), c \in \mathcal{C}\}$ as the pool of category-specific mappings associated to $\epsilon$. Here $\epsilon^{\perp}$ denotes the $\epsilon$-orthogonal reduction, \eg if $\epsilon$ is max-pooling along one dimension then $\epsilon^{\perp}$ is max-pooling along the other dimensions. That is, our pooling model is determined by the triplet
\begin{equation}
  (\varsigma) = (\epsilon, \pi, \{\theta_c\})\text{ , }\quad\pi\textrm{ is }\epsilon\text{-equivariant}
\end{equation}
and realized on a feature tensor $\io{x}$ by
\begin{eqnarray}
  \varsigma(\io{x},\{\theta_c\}) &=& \epsilon^{\perp}(\pi(\io{x},\theta_{c^*}))\label{eq:pooling.0}\\
  \textrm{where }c^* &=& \argmax_c\pi(\epsilon(\io{x}),\theta_c)\label{eq:pooling.1}
\end{eqnarray}

In our model we choose
\begin{eqnarray*}
  \epsilon(\io{x}) 
  &\leftarrow& \text{ spatial average pooling}\\
  \pi(\epsilon,\theta_c) 
  &\leftarrow& \text{ linear mapping}
\end{eqnarray*}
so $\varsigma(\io{x},\{\theta_c\})$ is a differentiable spatial mapping, \ie, we can use $\varsigma$ as a trainable attention model for $\io{x}$. This is related to class activation mapping~\cite{zhou15cnnlocalization} introduced for discriminative localization. Note however that, in contrast to~\cite{zhou15cnnlocalization} that uses strong supervision to train the selector directly, we leverage video-level annotation to implicitly learn an attention mechanism for video classification. Our formulation is also a generalization: other choices are possible for the reduction $\epsilon$, and the use of differentiable structured layers~\cite{sminchisescu15iccv} in this context are an interesting direction for future work.

To inflate attention in \ac{lsta}, we introduce a new state tensor $\io{a_t}$ of shape $N \times 1$. Its update rule is that of standard \ac{lstm} (Eq.~\ref{eq:lsta.2}) with gatings $(f_a,i_a,\io{s_t})$ and innovation $a$ computed from the pooled $\nu_a = \varsigma(\io{x_t},w_a)$ as input (Eq.~\ref{eq:lsta.1}). We compute the attention tensor $s$ using the hidden state $\io{s_t} \odot \tanh(\io{a_t})$ as residual (Eq.~\ref{eq:lsta.3}), followed by a softmax calibration. Eqs.~\ref{eq:lsta.4}-\ref{eq:lsta.7} implement the \ac{lsta} memory update based on the filtered input $s \odot \io{x_t}$, this is described next.

\subsection{Output Pooling}
\label{sec:pooling}

If we analyze standard \ac{lstm} Eq.~\ref{eq:lstm.1} with input $s\odot\io{x_t}$ instead of $\io{x_t}$, it becomes evident that $\io{o_{t-1}}$ (output gating) has on $\io{c_{t-1}}$ a same effect as $s$ (attention) has on $\io{x_t}$. Indeed, in Eq.~\ref{eq:lsta.4} the gatings and innovation are all computed from $[s \odot \io{x_t}, \io{o_{t-1}} \odot \tanh(\io{c_{t-1}})]$. We build upon this analogy to enhance the output gating capacity of \ac{lsta} and, consequently, its forget-update behavior of memory tracking.

We introduce attention pooling in the output gating update. Instead of computing $\io{o_t}$ as by Eq.~\ref{eq:lstm.1} we replace $s\odot\io{x_t}$ with $\nu_c\odot\io{c_t}$ to obtain update Eqs.~\ref{eq:lsta.6}-\ref{eq:lsta.7}, that is
\begin{eqnarray*}
  \sigma(W_o * [s\odot\io{x_t}, \io{o_{t-1}} \odot \tanh(\io{c_{t-1}})]) &\leftarrow& \parbox[t]{1.3cm}{standard gating}\\
  \begin{matrix*}[r]
    \sigma(W_o * [\nu_c\odot\io{c_t}, \io{o_{t-1}} \odot \tanh(\io{c_{t-1}})])\\[.05in]
    \textrm{with }\nu_c = \varsigma(\io{c_t}, w_c + w_o \epsilon(s \odot \io{x_t}))
  \end{matrix*}
  &\leftarrow& \parbox[t]{1.3cm}{output pooling}
\end{eqnarray*}

This choice is motivated as follows. We want to preserve the recursive nature of output gating, which is we keep right-concatenating $\io{o_{t-1}} \odot \tanh(\io{c_{t-1}})$ to obtain the $2N \times K$-shaped tensor to convolve and tanh point-wise. Since the new memory state $\io{c_t}$ is available at this stage, which already integrates $s \odot \io{x_t}$, we can use this for left-concatenating instead of the raw attention-pooled input tensor. This is similar to a peephole connection in the output gate~\cite{gers2000recurrent}. We can even produce a filtered version $\nu_c \odot \io{c_t}$ of it if we introduce a second attention pooling neuron for localizing the actual discriminative memory component of $\io{c_t}$, that is via $\nu_c$, Eq.~\ref{eq:lsta.6}. Note that $\io{c_t}$ integrates information from past memory updates by design, so localizing current activations is pretty much required here. Consequently, and in contrast to feature tensors $\io{x_t}$, the memory activations might not be well localized spatially. We thus use a slightly different version of Eq.~\ref{eq:pooling.0} for output pooling, we remove $\epsilon^{\perp}$ to obtain a full-rank $N \times K$-shaped attention tensor $\nu_c$. 

To further enhance active memory localization, we use $s \odot \io{x_t}$ to control the bias term of attention pooling, Eq.~\ref{eq:lsta.6}. We apply a reduction $\epsilon(s \odot \io{x_t})$ followed by a linear regression with learnable parameters $w_o$ to obtain the instance-specific bias $w_o \epsilon(s \odot \io{x_t})$ for activation mapping. Note that $\epsilon$ is the reduction associated to $\varsigma$ so this is consistent. We will use a similar idea in Sec.~\ref{sec:twostream.cross} for cross-modal fusion in two-stream architecture. Our ablation study in Sec.~\ref{sec:exp.ablation} confirms that this further coupling of $\io{c_t}$ with $\io{x_t}$ boosts the memory distillation in the \ac{lsta} recursion, and consequently its tracking capability, by a significant margin.

\section{Two Stream Architecture}
\label{sec:architecture}


In this section, we explain our network architecture for egocentric activity recognition incorporating the \ac{lsta} module of Sec.~\ref{sec:lsta}. Like the majority of the deep learning methods proposed for action recognition, we also follow the two stream architecture; one stream for encoding appearance information from RGB frames and the second stream for encoding motion information from optical flow stacks. 
\subsection{Attention on Appearance Stream}
\label{sec:app_stream}
	The network consists of a ResNet-34 pre-trained on imageNet for image recognition. We use the output of the last convolution layer of block \verb+conv5_3+ of ResNet-34 as the input of the \ac{lsta} module. From this frame level features, \ac{lsta} generates the attention map which is used to weight the input features. We select 512 as the depth of LSTA memory and all the gates use a kernel size of $3\times 3$. We use the internal state~($\io{c_t}$) for classification.
	
	We follow a two stage training. In the first stage, the classifier and the \ac{lsta} modules are trained while in the second stage, the convolutional layers in the final block (\verb+conv5_x+) and the FC layer of ResNet-34 along with the layers trained in stage 1 are trained. 
	
\subsection{Attention on Motion Stream}
\label{sec:motion_stream}
	We use a network trained on optical flow stacks for explicit motion encoding. For this, we use a ResNet-34 CNN. The network is first trained on action verbs (take, put, pour, open, etc.) using an optical flow stack of 5 frames. We average the weights in the input convolutional layer of an imagenet pre-trained network and replicate it 10 times to initialize the input layer. 
	This is analogous to the imageNet pre-training done on the appearance stream. The network is then trained for activity recognition as follows. We use the action-pretrained ResNet-34 FC weights as the parameter initialization of attention pooling (Eqs.~\ref{eq:pooling.0}-\ref{eq:pooling.1}) on \verb+conv5_3+ flow features. We use this attention map to weight the features for classification. Since the activities are temporally located in the videos and they are not sequential in nature, we take the optical flow corresponding to the five frames located in the temporal center of the videos.
\subsection{Cross-modal Fusion}
\label{sec:twostream.cross}
    Majority of the existing methods with two stream architecture perform a simple late fusion by averaging for combining the outputs from the appearance and motion streams \cite{simonyan2014two, TSN2016ECCV}. Feichtenhofer \etal \cite{feichtenhofer2016convolutional} propose a pooling strategy at the output of the final convolutional layer for improved fusion of the two streams. In \cite{feichtenhofer2016spatiotemporal} the authors observe that adding a residual connection from the motion stream to the appearance stream enables the network to improve the joint modeling of the information flowing through the two streams. Inspired by the aforementioned observations, we propose a novel cross-modal fusion strategy in the earlier layers of the network in order to facilitate the flow of information across the two modalities. 

    In the proposed cross-modal fusion approach, each stream is used to control the biases of the other as follows. To perform cross-modal fusion on the appearance stream, the flow feature from the \verb+conv5_3+ of the motion stream CNN is applied as bias to the gates of the LSTA layer. To perform cross-modal fusion on the motion stream instead, the sequence of features from the \verb+conv5_3+ of the RGB stream CNN are 3D convolved into a summary feature. We add a \ac{clstm} cell of memory size 512 in the motion stream as an embedding layer and use the RGB summary feature to control the bias of the \ac{clstm} gates. 
    
    
    In this way, each individual stream is made to influence the encoding of the other so that we have a flow of information between them deep inside the neural network. We then perform a late average fusion of the two individual streams' output to obtain the class scores. 


\section{Experiments and Results}
\label{sec:results}
\vspace{-0.1cm}
\subsection{Datasets}
\vspace{-0.1cm}
	We evaluate the proposed method on four standard first person activity recognition datasets namely, GTEA 61, GTEA 71, EGTEA Gaze+ and EPIC-KITCHENS. GTEA 61 and GTEA 71 are relatively small scale datasets with 61 and 71 activity classes respectively. EGTEA Gaze+ is a recently developed large scale dataset with approximately 10K samples having 106 activity classes. EPIC-KITCHENS dataset is the largest egocentric activities dataset available now. The dataset consists of more than 28K video samples with 125 verb and 352 noun classes.

\subsection{Experimental Settings}
	The appearance and motion networks are first trained separately followed by a combined training of the two stream cross-modal fusion network.	We train the networks for minimizing the cross-entropy loss. The appearance stream is trained for 200 epochs in stage 1 with a learning rate of 0.001 which is decayed after 25, 75 and 150 epochs at a rate of 0.1. In the second stage, the network is trained with a learning rate of 0.0001 for 100 epochs. The learning rate is decayed by 0.1 after 25 and 75 epochs. We use ADAM as the optimization algorithm. 25 frames uniformly sampled from the videos are used as input. The number of classes used in the output pooling ($w_c$ in \ref{sec:pooling}) is chosen as 100 for GTEA 61 and GTEA 71 datasets after empirical evaluation on the fixed split of GTEA 61. For EGTEA Gaze+ and EPIC-KITCHENS datasets, the value is scaled to 150 and 300 respectively, in accordance with the relative increase in the number of activity classes.

	For the pre-training of the motion stream on action classification task, we use a learning rate of 0.01 which is reduced by 0.5 after 75, 150, 250 and 500 epochs and is trained for 700 epochs. In the activity classification stage, we train the network for 500 epochs with a learning rate of 0.01. The learning rate is decayed after 50 and 100 epochs by 0.5. SGD algorithm is used for optimizing the parameter updates of the network.
	
	The two stream network is trained for 200 epochs for GTEA 61 and GTEA 71 datasets while EGTEA is trained till 100 epochs, with a learning rate of 0.01 using ADAM algorithm. Learning rate is reduced by 0.99 after each epoch. We use a batch size of 32 for all networks. We use random horizontal flipping and multi-scale corner cropping techniques proposed in \cite{TSN2016ECCV} during training and the center crop of the frame is used during inference.

	\subsection{Ablation Study}
	\begin{table}[t]\small
	\centering
	\begin{tabular}{|l|c|}
		\hline
		Ablation & Accuracy (\%) \\ \hline \hline
		Baseline & 51.72 \\ \hline
		Baseline + output pooling & 62.07\\ \hline
		Baseline + attention pooling & 66.38 \\ \hline
		Baseline + pooling & 68.1 \\ \hline
		\textbf{\ac{lsta}} & \textbf{74.14} \\ \hline \hline
		LSTA two stream late fusion  & 78.45 \\ \hline
		\textbf{LSTA two stream cross-modal fusion} & \textbf{79.31} \\ \hline
	\end{tabular}
	\vspace{-0.2cm}
	\caption{Ablation analysis on GTEA 61 fixed split.}
	\label{tab:ablation}\vspace{-0.3cm}
\end{table}

An extensive ablation analysis\footnote{Detailed analysis available in the supplementary document.} has been carried out, on the fixed split of GTEA 61 dataset, to determine the performance improvement obtained by each component of \ac{lsta}.
The results are shown in Tab.~\ref{tab:ablation},
which compares the performance of RGB and two stream networks on the top and bottom sections respectively.
We choose a network with vanilla \ac{clstm} as the baseline since \ac{lsta} without attention and output pooling converges to the standard \ac{clstm}. The baseline model results in an accuracy of $51.72\%$. We then analyze the impact of each of the contributions explained in Sec \ref{sec:lsta}. We first analyze the effect of output pooling on the baseline. By adding output pooling the performance is improved by $8\%$. We analyzed the classes that are improved by adding output pooling over the baseline model and observe that the major improvement is achieved by predicting the correct action classes. Output pooling enables the network to propagate a filtered a version of the memory which is localized on the most discriminative components.

Adding attention pooling to the baseline improves the performance by $14\%$. Attention pooling enables the network to identify the relevant  regions in the input frame and to maintain a history of the relevant regions seen in the past frames. This enables the network to have a smoother tracking of attentive regions. Detailed analysis show that attention pooling enables the network to correctly classify activities with multiple objects. It should be noted that this is equivalent to a network with two \ac{clstm}s, one for attention tracking and one for frame level feature tracking. 

Incorporating both attention and output pooling to the baseline results in a gain of $16\%$. By analyzing the top improved classes, we found that the model has increased its capacity to correctly classify both actions and objects. By adding bias control, as explained in Sec.~\ref{sec:lsta}, we obtain the proposed \ac{lsta} model and gains an additional improvement of $6\%$ in recognition accuracy.

Compared to the network with the vanilla \ac{clstm}, \ac{lsta} achieves an improvement of $22\%$. From the previous analyses we have seen the importance of attention pooling and output pooling present in \ac{lsta}. This enables the network to focus on encoding the features more relevant for the concrete classification task. Detailed analysis shows \ac{clstm} confuses with both activities involving same action with different objects as well as activities consisting of different action with same objects. With the attention mechanism, \ac{lsta} weights the most discriminant features, thereby allowing the network to distinguish between the different activity classes.

\begin{table}[t]\small
	\centering
	\begin{tabular}{|l|c|}
		\hline
		Method & Accuracy (\%) \\ \hline \hline
        eleGAtt~\cite{attention_eccv18} & 59.48 \\ \hline
		ego-rnn~\cite{sudhakaran2018attention} & 63.79 \\ \hline
		\textbf{\ac{lsta}} & \textbf{74.14} \\ \hline \hline
		ego-rnn two stream~\cite{sudhakaran2018attention} & 77.59 \\ \hline
		\textbf{\ac{lsta} two stream} & \textbf{79.31} \\ \hline
	\end{tabular}
\vspace{-0.2cm}
	\caption{Comparative analysis on GTEA 61 fixed split.}
	\label{tab:comp}\vspace{-0.4cm}
\end{table}

We also evaluated the performance improvement achieved by applying attention to the motion stream. The baseline is a ResNet-34 pre-trained on actions followed by training for activities. We obtained an accuracy of $40.52\%$ for the network with attention compared to the $36.21\%$ of the baseline. Fig.~\ref{fig:att_map} (fourth row) visualizes the attention map generated by the network. For visualization, we overlay the resized attention map on the RGB frames corresponding to the optical flow stack used as input. From the figure, it can be seen that the network generates the attention map around/near the hands, where the discriminant motion is occurring, thereby enabling the network to recognize the activity undertaken by the user. It can also be seen that the attention maps generated by the appearance stream and the flow stream are complementary to each other; appearance stream focuses on the object regions while the motion stream focuses on hand regions. We also analyzed the classes where the network with attention performs better compared to the standard flow network and found that the network with attention is able to recognize actions better than the standard network. This is because the attention mechanism enables the network to focus on regions where motion is occurring in the frame.

Next we compare the performance of the cross-modal fusion technique explained in Sec.~\ref{sec:twostream.cross} over traditional late fusion two stream approach. The cross-modal fusion approach improves by $1\%$ over late fusion. Analysis shows that the cross-modal fusion approach is able to correctly identify activities with same objects. The fifth and sixth rows of Fig.~\ref{fig:att_map} visualize the attention maps generated after cross-modal fusion training. It can be seen that the motion stream attention expands to regions containing objects. This validates the effect of cross-modal fusion where the two networks are made to interact deep inside the network.

\subsection{Comparative Analysis}

\begin{figure*}[t]
		\centering      
		\raisebox{.19in}{\rotatebox[origin=t]{90}{Input}}
        \begin{subfigure}[b]{0.4\textwidth}
			\caption*{Close chocolate}
\vspace{-0.18cm}
			\includegraphics[scale=0.17]{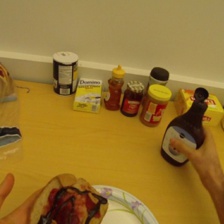}
			\includegraphics[scale=0.17]{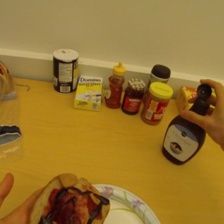}
			\includegraphics[scale=0.17]{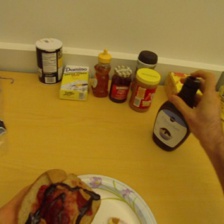}
			\includegraphics[scale=0.17]{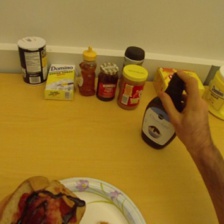}
			\includegraphics[scale=0.17]{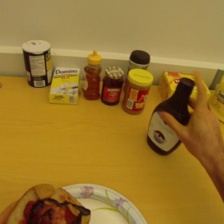}
			\end{subfigure} \hskip 5mm
	        \begin{subfigure}[b]{0.4\textwidth}
			\caption*{Scoop coffee}
\vspace{-0.18cm}
			\includegraphics[scale=0.17]{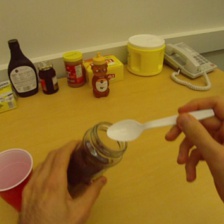}
			\includegraphics[scale=0.17]{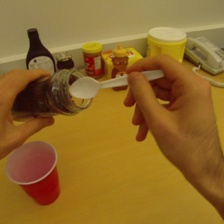}
			\includegraphics[scale=0.17]{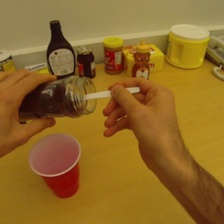}
			\includegraphics[scale=0.17]{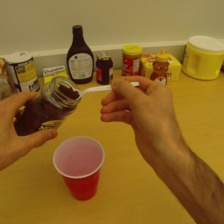}
			\includegraphics[scale=0.17]{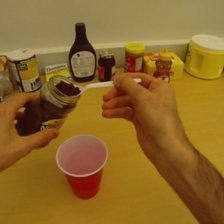}
		\end{subfigure}\\
	\ 	\raisebox{.19in}{\rotatebox[origin=t]{90}{ego-rnn}}
        \begin{subfigure}[b]{0.4\textwidth}
			\includegraphics[scale=0.17]{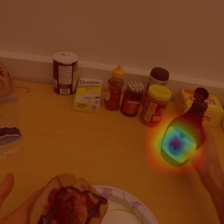}
			\includegraphics[scale=0.17]{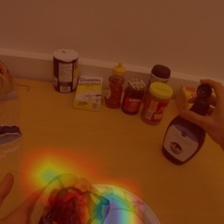}
			\includegraphics[scale=0.17]{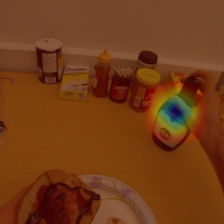}
			\includegraphics[scale=0.17]{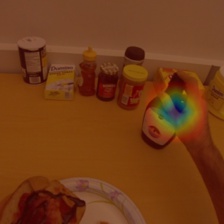}
			\includegraphics[scale=0.17]{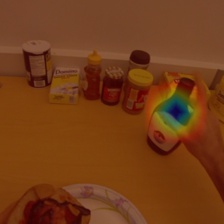}
			\end{subfigure} \hskip 5mm
	        \begin{subfigure}[b]{0.4\textwidth}
			\includegraphics[scale=0.17]{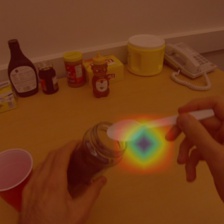}
			\includegraphics[scale=0.17]{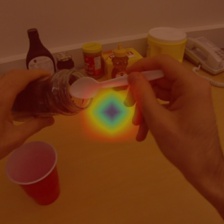}
			\includegraphics[scale=0.17]{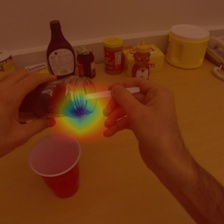}
			\includegraphics[scale=0.17]{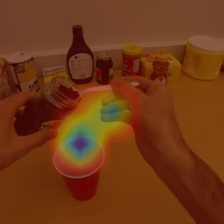}
			\includegraphics[scale=0.17]{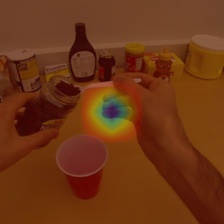}
		\end{subfigure}\\ 
	\	\raisebox{.19in}{\rotatebox[origin=t]{90}{LSTA}}
        \begin{subfigure}[b]{0.4\textwidth}
			\includegraphics[scale=0.17]{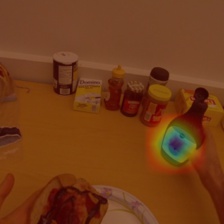}
			\includegraphics[scale=0.17]{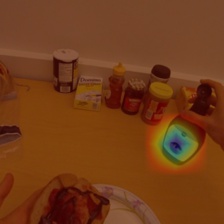}
			\includegraphics[scale=0.17]{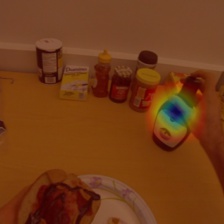}
			\includegraphics[scale=0.17]{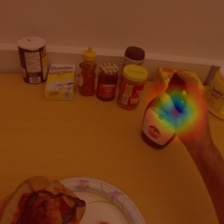}
			\includegraphics[scale=0.17]{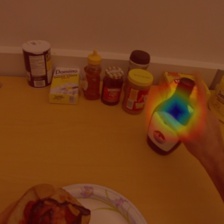}
			\end{subfigure} \hskip 5mm
	        \begin{subfigure}[b]{0.4\textwidth}
			\includegraphics[scale=0.17]{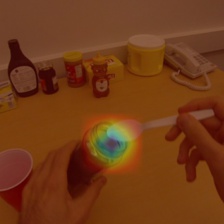}
			\includegraphics[scale=0.17]{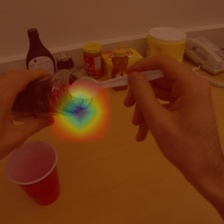}
			\includegraphics[scale=0.17]{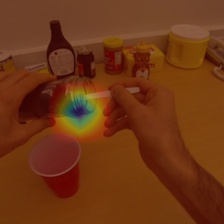}
			\includegraphics[scale=0.17]{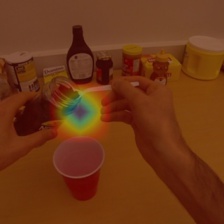}
			\includegraphics[scale=0.17]{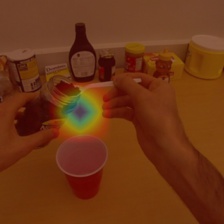}
		\end{subfigure}\\
	\	\raisebox{.19in}{\rotatebox[origin=t]{90}{Flow}}
       \begin{subfigure}[b]{0.4\textwidth}
		    \includegraphics[scale=0.17]{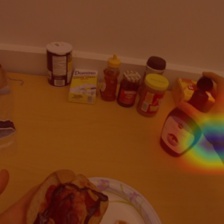}
			\includegraphics[scale=0.17]{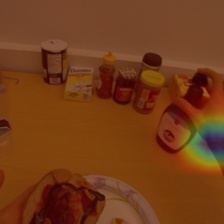}
			 \includegraphics[scale=0.17]{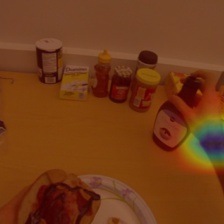}
			\includegraphics[scale=0.17]{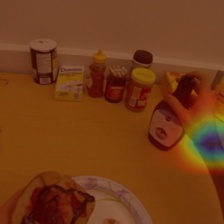}
			\includegraphics[scale=0.17]{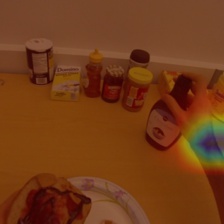}
			\end{subfigure} \hskip 5mm
	        \begin{subfigure}[b]{0.4\textwidth}
			\includegraphics[scale=0.17]{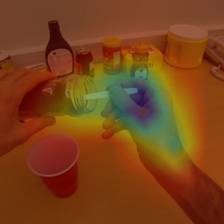}
			\includegraphics[scale=0.17]{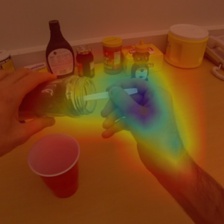}
			\includegraphics[scale=0.17]{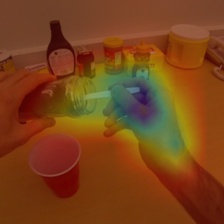}
			\includegraphics[scale=0.17]{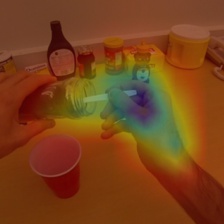}
			\includegraphics[scale=0.17]{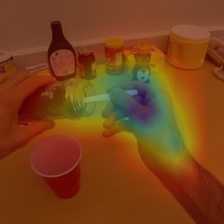}
		\end{subfigure}\\
	\	\raisebox{.19in}{\rotatebox[origin=t]{90}{LSTA$^*$}}
        \begin{subfigure}[b]{0.4\textwidth}
			\includegraphics[scale=0.17]{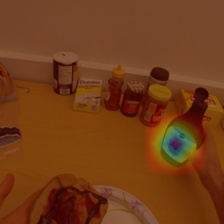}
			\includegraphics[scale=0.17]{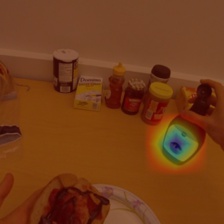}
			\includegraphics[scale=0.17]{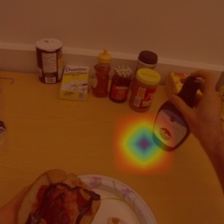}
			\includegraphics[scale=0.17]{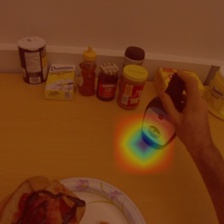}
			\includegraphics[scale=0.17]{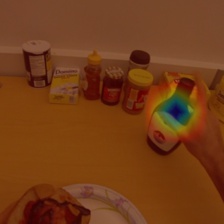}
			\end{subfigure} \hskip 5mm
	        \begin{subfigure}[b]{0.4\textwidth}
			\includegraphics[scale=0.17]{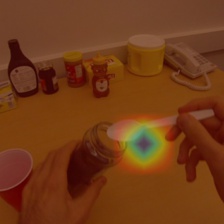}
			\includegraphics[scale=0.17]{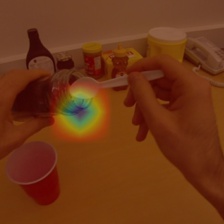}
			\includegraphics[scale=0.17]{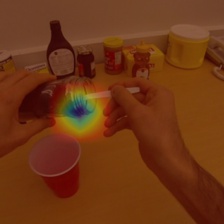}
			\includegraphics[scale=0.17]{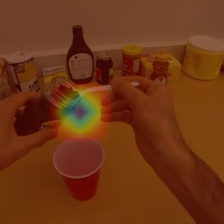}
			\includegraphics[scale=0.17]{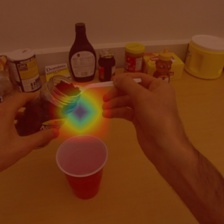}
		\end{subfigure} \\
	 \	\raisebox{.19in}{\rotatebox[origin=t]{90}{Flow$^*$}}
       \begin{subfigure}[b]{0.4\textwidth}
			\includegraphics[scale=0.17]{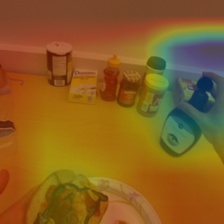}
			\includegraphics[scale=0.17]{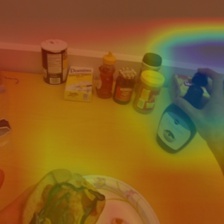}
			\includegraphics[scale=0.17]{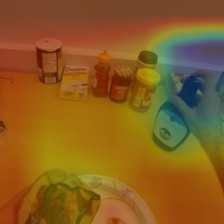}
			\includegraphics[scale=0.17]{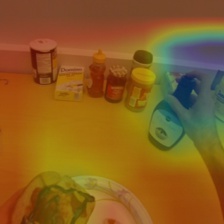}
			\includegraphics[scale=0.17]{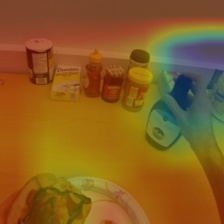}
			\end{subfigure} \hskip 5mm
	        \begin{subfigure}[b]{0.4\textwidth}
			\includegraphics[scale=0.17]{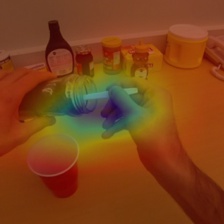}
			\includegraphics[scale=0.17]{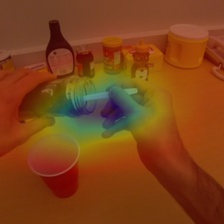}
			\includegraphics[scale=0.17]{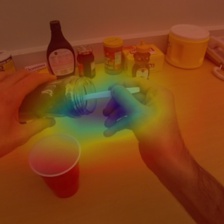}
			\includegraphics[scale=0.17]{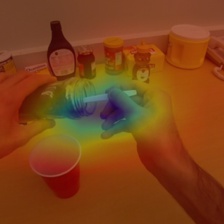}
			\includegraphics[scale=0.17]{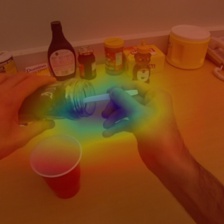}
		\end{subfigure}
        \caption{Attention maps generated by ego-rnn (second row) and \ac{lsta} (third) for two video sequences. We show the 5 frames that are uniformly sampled from the 25 frames used as input to the corresponding networks. Fourth row shows the attention map generated by the motion stream. Fifth and sixth rows show the attention map generated by the appearance and flow streams after two stream cross-modal training. For flow, we visualize the attention map on the five frames corresponding to the optical flow stack given as input. ($^*$: Attention map obtained after two stream cross-modal fusion training).}
		\label{fig:att_map}
	\end{figure*}

In this section, we compare the performance of \ac{lsta} over two closely related methods, namely, eleGAtt~\cite{attention_eccv18} and ego-rnn \cite{sudhakaran2018attention}. Results are shown in Tab.~\ref{tab:comp}. EleGAtt is an attention mechanism which can be applied to any generic RNN using its hidden state for generating the attention map. We evaluated eleGAtt on \ac{lstm}, consisting of 512 hidden units, with the same training setting as \ac{lsta} for fair comparison. EleGAtt learns a single weight matrix for generating the attention map irrespective of the input whereas \ac{lsta} generates the attention map from a pool of weights which are selected in a top-down manner based on input. This enables the selection of a proper attention map for each input activity class. This leads to a performance gain of $13\%$ over eleGAtt. Analyzing the classes with the highest improvement by \ac{lsta} compared to eleGAtt reveals that~eleGAtt fails in identifying the object while correctly classifying the action. Ego-rnn \cite{sudhakaran2018attention} derives an attention map generated from class activation map to weight the discriminant regions in the image which are then applied to a \ac{clstm} cell for temporal encoding. It generates a per frame attention map which has no dependency on the information present in the previous frames. This can result in selecting different objects in adjacent frames. On the contrary, \ac{lsta} uses an attention memory to track the previous attention maps enabling their smooth tracking. This results in a $10\%$ improvement obtained by \ac{lsta} over ego-rnn. Detailed analysis on the classification results show that ego-rnn struggles to classify activities involving multiple objects. Since the attention map generated in each frame is independent of the previous frames, the network fails to track previously activated regions, thereby resulting in wrong predictions. This is further illustrated by visualizing the attention maps produced by ego-rnn and \ac{lsta} in Fig.~\ref{fig:att_map}. From the figure, one can see that ego-rnn (second row) fails to identify the relevant object in the case of close chocolate example and it failed to track the object in the final frames in the case of the scoop coffee example. \ac{lsta} with cross-modal fusion performs $2\%$ better than  ego-rnn two stream. 


\vspace{-0.1cm}
\label{subsec:spat_att}

	\label{sec:exp.ablation}

	\subsection{State-of-the-art comparison}
	
	Our approach is compared against the state-of-the-art methods on Tab. \ref{tab:res_table}. The methods listed in the first section of the table uses strong supervision signals such as gaze \cite{li2015delving, li2018eye}, hand segmentation \cite{ma2016deeper} or object bounding boxes \cite{ma2016deeper} during the training stage. Two stream \cite{simonyan2014two}, I3D \cite{carreira2017quo} and TSN \cite{TSN2016ECCV} are methods proposed for action recognition from third person videos while all other methods except eleGAtt \cite{attention_eccv18} are proposed for first-person activity recognition. eleGAtt \cite{attention_eccv18} is proposed as a generic method for incorporating attention mechanism to any RNN module. From the table, we can see that the proposed method outperforms all the existing methods for egocentric activity recognition. 

	

	\begin{table}[t]\small
		\begin{center}\setlength{\tabcolsep}{4pt} 
			\begin{tabular}{|l|c|c|c|c|}
				\hline
	Methods & GTEA61$^*$ & GTEA61 & GTEA71 & EGTEA\\
			\hline
				Li \etal \cite{li2015delving}$^{**}$ & 66.8 & 64 & 62.1 & 46.5 \\
				\hline
				Ma \etal \cite{ma2016deeper}$^{**}$ & 75.08 & 73.02 & 73.24 & - \\
				\hline
				Li \etal \cite{li2018eye}$^{**}$ & - & - & - & 53.3\\
				\hline
				Two stream \cite{simonyan2014two} & 57.64 & 51.58 & 49.65 & 41.84 \\
				\hline
				I3D \cite{carreira2017quo} & - & - & - & 51.68 \\
				\hline
				TSN \cite{TSN2016ECCV}  & 67.76 & 69.33 & 67.23 & 55.93 \\ 
				\hline
				eleGAtt \cite{attention_eccv18} & 59.48 & 66.77 & 60.83 & 57.01 \\
				\hline
				ego-rnn \cite{sudhakaran2018attention} & 77.59 & 79 & 77 & 60.76 \\
				\hline
				\ac{lsta}-RGB & 74.14 & 71.32 & 66.16
 & 57.94 \\
				\hline
				\textbf{\ac{lsta}} & \textbf{79.31} & \textbf{80.01} & \textbf{78.14} & \textbf{61.86} \\
				\hline
			\end{tabular}
		\end{center}
		\vspace{-0.6cm}
		\caption{Comparison with state-of-the-art methods on popular egocentric datasets, we report recognition accuracy in \%. ($^*$: fixed split; $^{**}$: trained with strong supervision).}
		\label{tab:res_table}\vspace{-0.5cm}
	\end{table}

In EPIC-KITCHENS dataset, the labels are provided in the form of \verb+verb+ and \verb+noun+, which are combined to form an activity class. The fact that not all combinations of verbs and nouns are feasible and that not all test classes might have a representative training sample make it a challenging problem. We train the network for multi-task classification with \verb+verb+, \verb+noun+ and \verb+activity+ supervision. We use \verb+activity+ classifier activations to control the bias of \verb+verb+ and \verb+noun+ classifiers. 
The dataset provides two evaluation settings, seen kitchens (S1) and unseen kitchens (S2). We obtained an accuracy of $30.16\%$ (S1) and $15.88\%$ (S2) using RGB frames. The best performing baseline is a two stream TSN that achieves $20.54\%$ (S1) and $10.89\%$ (S2)~\cite{Damen_2018_ECCV}. Our model is particularly strong on \verb+verb+ prediction (58\%) where we gain +10\% points over TSN. \verb+verb+ in this context is typically describing actions that develop into an activity over time, confirming once more \ac{lsta} efficiently learns encoding of sequences with localized patterns. 

\section{Conclusion}
\label{sec:conclusion}

We presented \ac{lsta} that extends \ac{lstm} with two core features: 1) attention pooling that spatially filters the input sequence and 2) output pooling that exposes a distilled view of the memory at each iteration. As shown in a detailed ablation study, both contributions are essential for a smooth and focused tracking of a latent representation of the video to achieve superior performance in classification tasks where the discriminative features can be localized spatially. We demonstrate its practical benefits for egocentric activity recognition with a two stream CNN-LSTA architecture featuring a novel cross-modal fusion and we achieve state-of-the-art accuracy on four standard benchmarks. 
\vskip 1.5mm

{\setlength{\parindent}{0cm}\small{\textbf{Acknowledgements:}
This work has been partially supported by the Spanish project TIN2016-74946-P (MINECO/FEDER, UE), CERCA Programme / Generalitat de Catalunya and ICREA under the ICREA Academia programme. We gratefully acknowledge the support of NVIDIA Corporation with the donation of GPUs used for this research.}}

{\small
\bibliographystyle{ieee}
\bibliography{lsta}

\begin{thebibliography}{10}\itemsep=-1pt

\bibitem{anderson2018bottom}
P. Anderson, X. He, C. Buehler, D. Teney, M. Johnson, S. Gould, and L. Zhang.
\newblock Bottom-up and top-down attention for image captioning and visual
  question answering.
\newblock In {\em Proc. CVPR}, 2018.

\bibitem{cao2017egocentric}
C. Cao, Y. Zhang, Y. Wu, H. Lu, and J. Cheng.
\newblock Egocentric gesture recognition using recurrent 3d convolutional
  neural networks with spatiotemporal transformer modules.
\newblock In {\em Proc. ICCV}, 2017.

\bibitem{carreira2017quo}
J. Carreira and A. Zisserman.
\newblock Quo vadis, action recognition? a new model and the kinetics dataset.
\newblock In {\em Proc. CVPR}, 2017.

\bibitem{Damen_2018_ECCV}
D. Damen, H. Doughty, G.M. Farinella, S. Fidler, A. Furnari, E. Kazakos, D.
  Moltisanti, J. Munro, T. Perrett, W. Price, and M. Wray.
\newblock Scaling egocentric vision: The epic-kitchens dataset.
\newblock In {\em Proc. ECCV}, 2018.

\bibitem{du2018recurrent}
W. Du, Y. Wang, and Y. Qiao.
\newblock Recurrent spatial-temporal attention network for action recognition
  in videos.
\newblock {\em IEEE Transactions on Image Processing}, 27(3):1347--1360, 2018.

\bibitem{feichtenhofer2016spatiotemporal}
C. Feichtenhofer, A. Pinz, and R. Wildes.
\newblock Spatiotemporal residual networks for video action recognition.
\newblock In {\em Proc. NIPS}, 2016.

\bibitem{feichtenhofer2016convolutional}
C. Feichtenhofer, A. Pinz, and A. Zisserman.
\newblock Convolutional two-stream network fusion for video action recognition.
\newblock In {\em Proc. CVPR}, 2016.

\bibitem{gers2000recurrent}
F.A. Gers and J. Schmidhuber.
\newblock Recurrent nets that time and count.
\newblock In {\em Proceedings of the IEEE-INNS-ENNS International Joint
  Conference on Neural Networks(IJCNN)}, 2000.

\bibitem{gers2000learning}
F.A. Gers, J. Schmidhuber, and F. Cummins.
\newblock {Learning to Forget: Continual Prediction with LSTM}.
\newblock {\em Neural Computation}, 12(10):2451--2471, 2000.

\bibitem{girdhar2017attentional}
R. Girdhar and D. Ramanan.
\newblock Attentional pooling for action recognition.
\newblock In {\em Proc. NIPS}, 2017.

\bibitem{he2017mask}
K. He, G. Gkioxari, P. Doll{\'a}r, and R. Girshick.
\newblock Mask r-cnn.
\newblock In {\em Proc. ICCV}, 2017.

\bibitem{he16residual}
K. He, X. Zhang, S. Ren, and J. Sun.
\newblock {Deep Residual Learning for Image Recognition}.
\newblock In {\em Proc. CVPR}, 2016.

\bibitem{Hochreiter:1997:LSM:1246443.1246450}
Sepp Hochreiter and J\"{u}rgen Schmidhuber.
\newblock Long short-term memory.
\newblock {\em Neural Comput.}, 9(8):1735--1780, 1997.

\bibitem{sminchisescu15iccv}
C. Ionescu, O. Vantzos, and C. Sminchisescu.
\newblock {Matrix Backpropagation for Deep Networks with Structured Layers}.
\newblock In {\em Proc. CVPR}, 2015.

\bibitem{li2018eye}
Y. Li, M. Liu, and J.M. Rehg.
\newblock In the eye of beholder: Joint learning of gaze and actions in first
  person video.
\newblock In {\em Proc. ECCV}, 2018.

\bibitem{li2015delving}
Y. Li, Z. Ye, and J.M Rehg.
\newblock {Delving into Egocentric Actions}.
\newblock In {\em Proc. CVPR}, 2015.

\bibitem{li2018videolstm}
Z. Li, K. Gavrilyuk, E. Gavves, M. Jain, and C.G.M. Snoek.
\newblock Videolstm convolves, attends and flows for action recognition.
\newblock {\em Computer Vision and Image Understanding}, 166:41--50, 2018.

\bibitem{liang2018focal}
J. Liang, L. Jiang, L. Cao, L. Li, and A. Hauptmann.
\newblock Focal visual-text attention for visual question answering.
\newblock In {\em Proc. CVPR}, 2018.

\bibitem{liu2016ssd}
W. Liu, D. Anguelov, D. Erhan, C. Szegedy, S. Reed, C. Fu, and A.C. Berg.
\newblock Ssd: Single shot multibox detector.
\newblock In {\em Proc. ECCV}, 2016.

\bibitem{ma2017attend}
C. Ma, A. Kadav, I. Melvin, Z. Kira, G. AlRegib, and H.P. Graf.
\newblock Attend and interact: Higher-order object interactions for video
  understanding.
\newblock In {\em Proc. CVPR}, 2018.

\bibitem{ma2016deeper}
M. Ma, H. Fan, and K.M. Kitani.
\newblock Going deeper into first-person activity recognition.
\newblock In {\em Proc. CVPR}, 2016.

\bibitem{nguyen2018improved}
D. Nguyen and T. Okatani.
\newblock Improved fusion of visual and language representations by dense
  symmetric co-attention for visual question answering.
\newblock In {\em Proc. CVPR}, 2018.

\bibitem{piergiovanni2017learning}
A. Piergiovanni, C. Fan, and M.S. Ryoo.
\newblock Learning latent sub-events in activity videos using temporal
  attention filters.
\newblock In {\em AAAI Conference on Artificial Intelligence}, 2017.

\bibitem{ryoo2015pooled}
M.S. Ryoo, B. Rothrock, and L. Matthies.
\newblock Pooled motion features for first-person videos.
\newblock In {\em Proc. CVPR}, 2015.

\bibitem{sharma2015action}
S. Sharma, R. Kiros, and R. Salakhutdinov.
\newblock Action recognition using visual attention.
\newblock In {\em Proc. ICLRW}, 2015.

\bibitem{shen2018egocentric}
Y. Shen, B. Ni, Z. Li, and N. Zhuang.
\newblock Egocentric activity prediction via event modulated attention.
\newblock In {\em Proc. ECCV}, 2018.

\bibitem{shi15convlstm}
X. Shi, Z. Chen, H. Wang, D. Yeung, W. Wong, and W. Woo.
\newblock {Convolutional LSTM Network: A Machine Learning Approach for
  Precipitation Nowcasting}.
\newblock In {\em Proc. NIPS}, 2015.

\bibitem{sigurdsson2018actor}
G. Sigurdsson, A. Gupta, C. Schmid, A. Farhadi, and K. Alahari.
\newblock Actor and observer: Joint modeling of first and third-person videos.
\newblock In {\em Proc. CVPR}, 2018.

\bibitem{simonyan2014two}
K. Simonyan and A. Zisserman.
\newblock {Two-Stream Convolutional Networks for Action Recognition in Videos}.
\newblock In {\em Proc. NIPS}, 2014.

\bibitem{singh2016first}
S. Singh, C. Arora, and CV Jawahar.
\newblock First person action recognition using deep learned descriptors.
\newblock In {\em Proc. CVPR}, 2016.

\bibitem{sudhakaran2017convolutional}
S. Sudhakaran and O. Lanz.
\newblock Convolutional long short-term memory networks for recognizing first
  person interactions.
\newblock In {\em Proc. ICCVW}, 2017.

\bibitem{sudhakaran2018attention}
S. Sudhakaran and O. Lanz.
\newblock Attention is all we need: Nailing down object-centric attention for
  egocentric activity recognition.
\newblock In {\em Proc. BMVC}, 2018.

\bibitem{sudhakaran2018top}
S. Sudhakaran and O. Lanz.
\newblock Top-down attention recurrent vlad encoding for action recognition in
  videos.
\newblock In {\em 17th International Conference of the Italian Association for
  Artificial Intelligence}, 2018.

\bibitem{tang2017action}
Y. Tang, Y. Tian, J. Lu, J. Feng, and J. Zhou.
\newblock Action recognition in rgb-d egocentric videos.
\newblock In {\em Proc. ICIP}, 2017.

\bibitem{tang2018multi}
Y. Tang, Z. Wang, J. Lu, J. Feng, and J. Zhou.
\newblock Multi-stream deep neural networks for rgb-d egocentric action
  recognition.
\newblock {\em IEEE Transactions on Circuits and Systems for Video Technology},
  2018.

\bibitem{verma2018making}
S. Verma, P. Nagar, D. Gupta, and C. Arora.
\newblock Making third person techniques recognize first-person actions in
  egocentric videos.
\newblock In {\em Proc. ICIP}, 2018.

\bibitem{wang2018bidirectional}
J. Wang, W. Jiang, L. Ma, W. Liu, and Y. Xu.
\newblock Bidirectional attentive fusion with context gating for dense video
  captioning.
\newblock In {\em Proc. CVPR}, 2018.

\bibitem{TSN2016ECCV}
L. Wang, Y. Xiong, Z. Wang, Y. Qiao, D. Lin, X. Tang, and L. {Van Gool}.
\newblock {Temporal Segment Networks: Towards Good Practices for Deep Action
  Recognition}.
\newblock In {\em Proc. ECCV}, 2016.

\bibitem{xie2017aggregated}
S. Xie, R. Girshick, P. Doll{\'a}r, Z. Tu, and K. He.
\newblock Aggregated residual transformations for deep neural networks.
\newblock In {\em Proc. CVPR}, 2017.

\bibitem{zaki2017modeling}
H.F.M. Zaki, F. Shafait, and A.S. Mian.
\newblock Modeling sub-event dynamics in first-person action recognition.
\newblock In {\em Proc. CVPR}, 2017.

\bibitem{attention_eccv18}
P. Zhang, J. Xue, C. Lan, W. Zeng, Z. Gao, and N. Zheng.
\newblock Adding attentiveness to the neurons in recurrent neural networks.
\newblock In {\em Proc. ECCV}, 2018.

\bibitem{zhou15cnnlocalization}
B. Zhou, A. Khosla, Lapedriza. A., A. Oliva, and A. Torralba.
\newblock {Learning Deep Features for Discriminative Localization}.
\newblock In {\em Proc. CVPR}, 2016.

\bibitem{zhou2016cascaded}
Y. Zhou, B. Ni, R. Hong, X. Yang, and Q. Tian.
\newblock Cascaded interactional targeting network for egocentric video
  analysis.
\newblock In {\em Proc. CVPR}, 2016.

\end{thebibliography}
}

\section*{Appendix}

	\section{Ablation Analysis}
\begin{figure*}[t]
		\centering
		\begin{subfigure}[b]{0.45\textwidth}
			\includegraphics[scale=0.53]{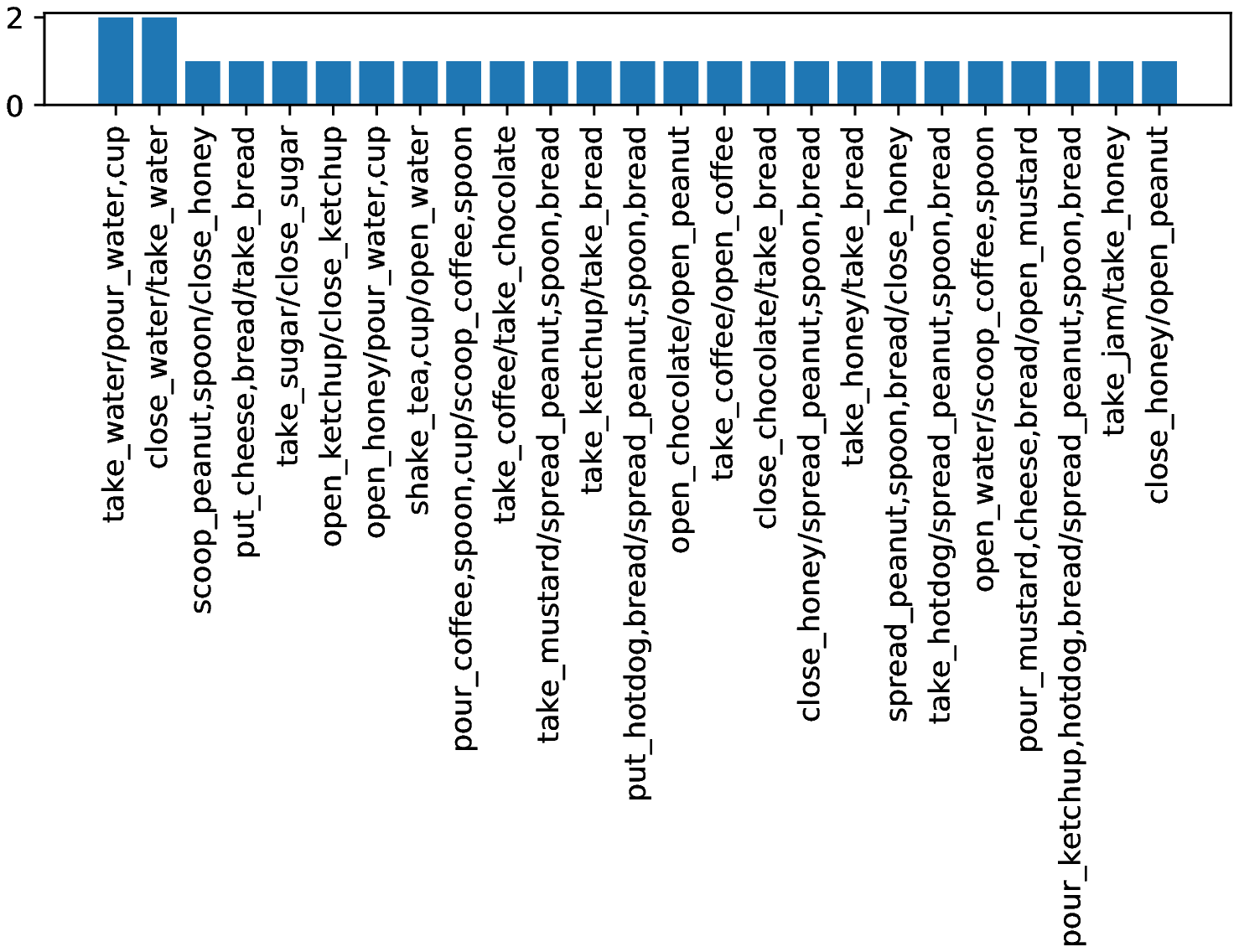}
			\caption{}
			\label{fig:base+outPoolingImp}
		\end{subfigure} \hspace{4mm}
		\begin{subfigure}[b]{0.45\textwidth}
			\includegraphics[scale=0.53]{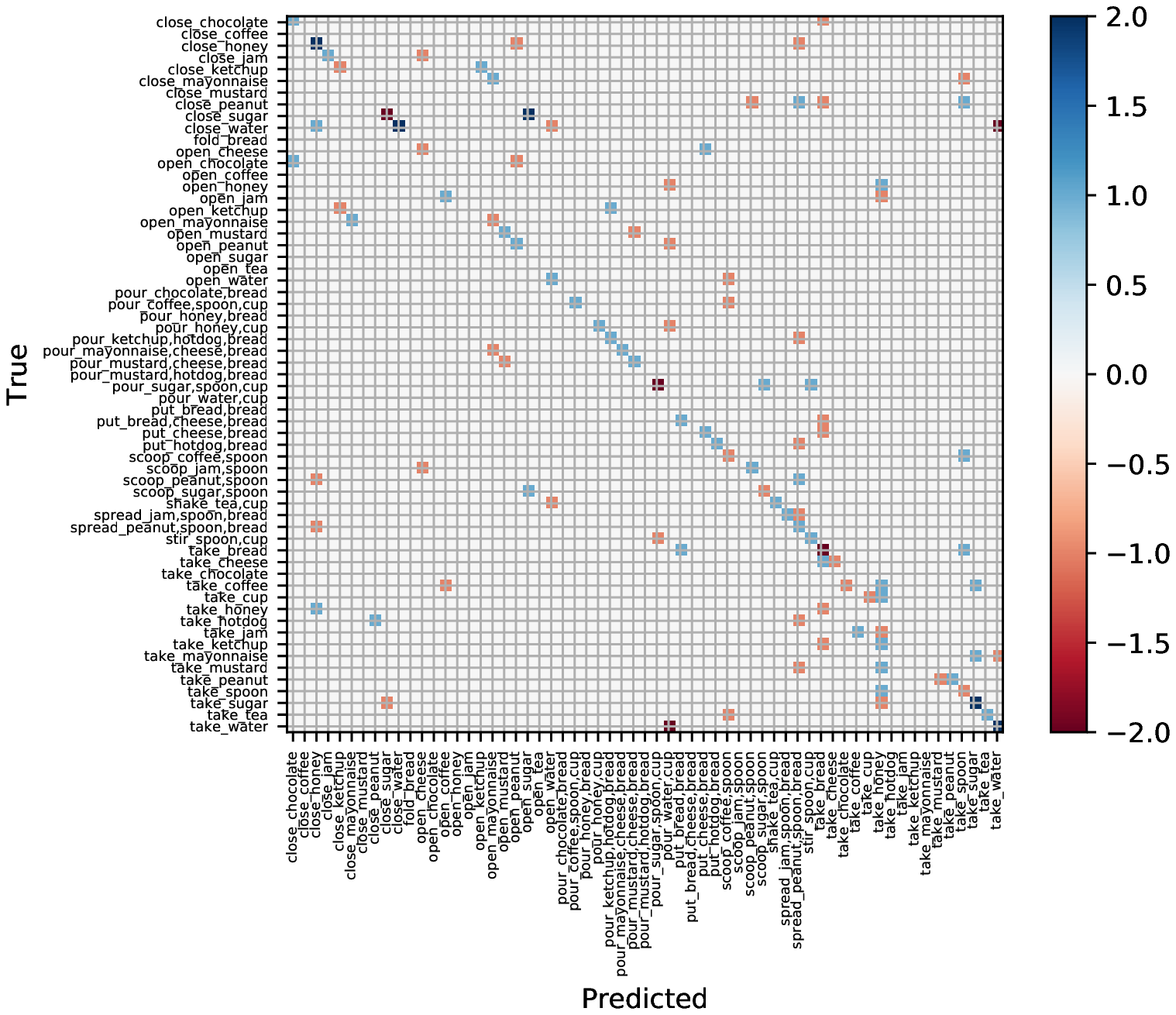}
			\label{fig:base+outPoolingConf}
			\caption{}
		\end{subfigure}
		\caption{(a) Most improvement categories by adding output pooling to the baseline on GTEA 61 fixed split. X axis labels are in the format true label (baseline + output pooling)/predicted label (baseline). Y axis shows the number of corrected samples for each class. (b) shows the difference of confusion matrices.}
		\label{fig:base+outPooling}
	\end{figure*}

\begin{figure*}[t]
		\centering
		\begin{subfigure}[b]{0.45\textwidth}
			\includegraphics[scale=0.53]{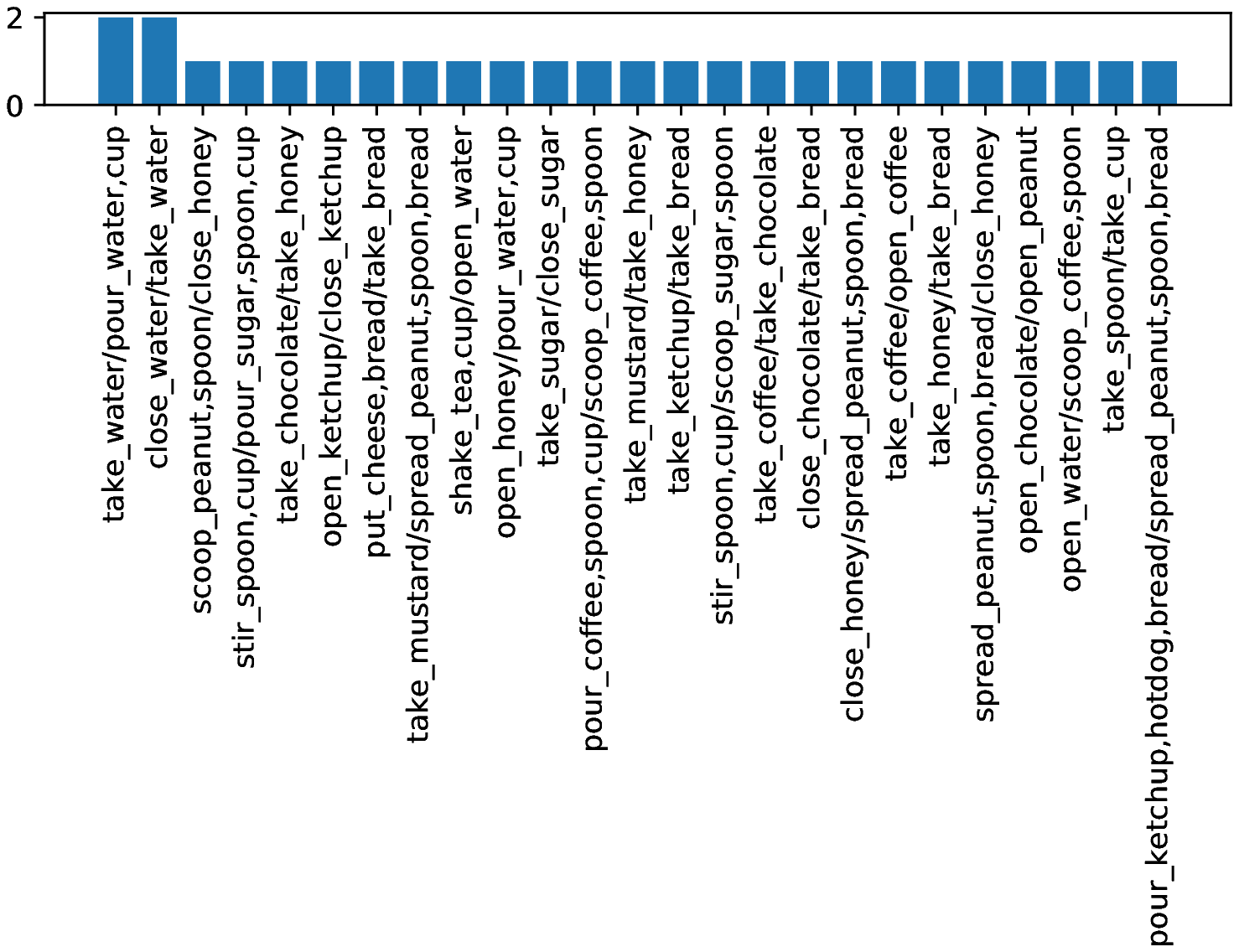}
			\label{fig:base+attPoolingImp}
			\caption{}
		\end{subfigure} \hspace{4mm}
		\begin{subfigure}[b]{0.45\textwidth}
			\includegraphics[scale=0.53]{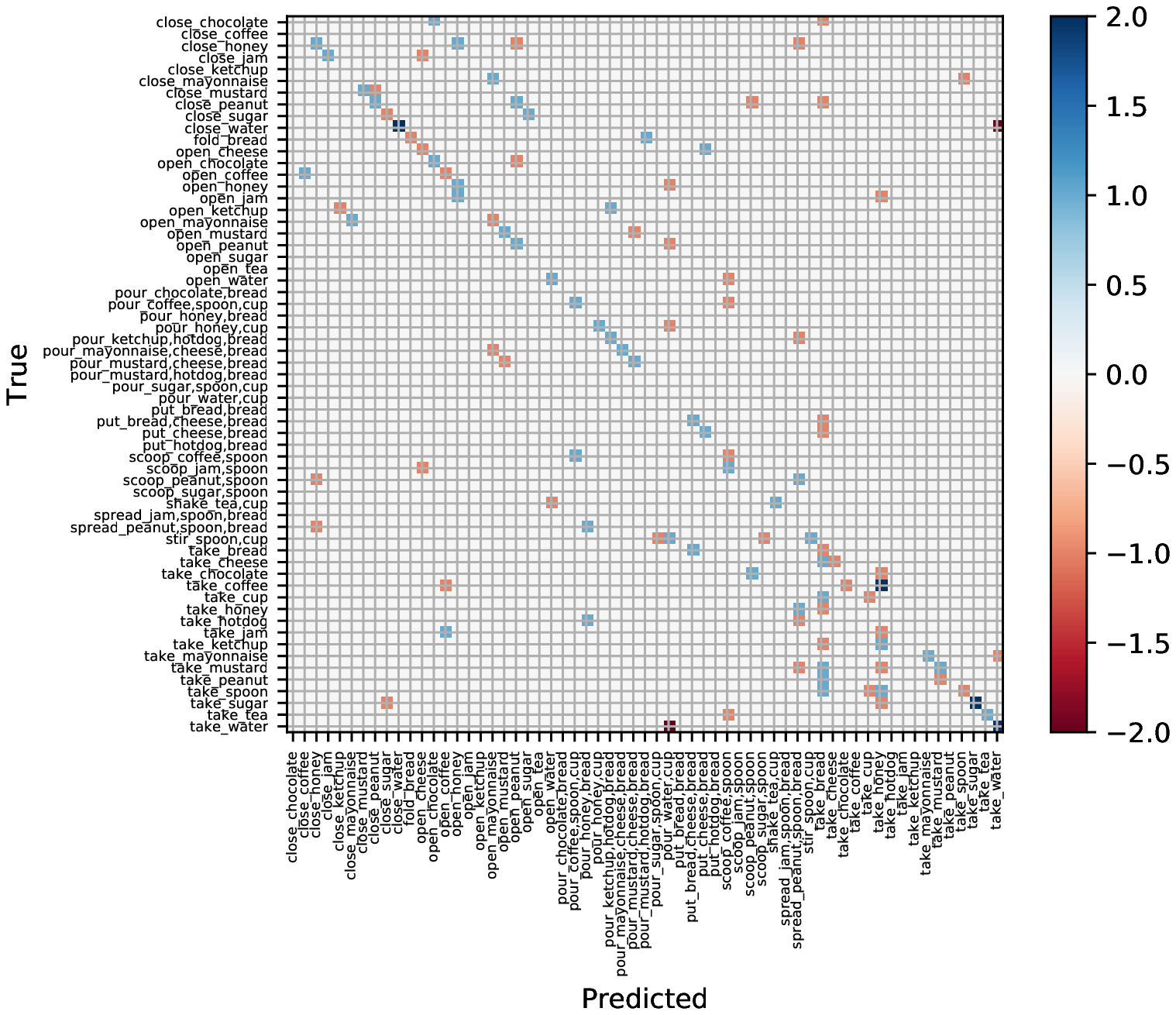}
			\label{fig:base+attPoolingConf}
			\caption{}
		\end{subfigure}
		\caption{(a) Most improvement categories by adding attention pooling to the baseline on GTEA 61 fixed split. X axis labels are in the format true label (baseline + attention pooling)/predicted label (baseline). Y axis shows the number of corrected samples for each class. (b) shows the difference of confusion matrices.}
		\label{fig:base+attPooling}
	\end{figure*}

\begin{figure*}[t]
		\centering
		\begin{subfigure}[b]{0.45\textwidth}
			\includegraphics[scale=0.53]{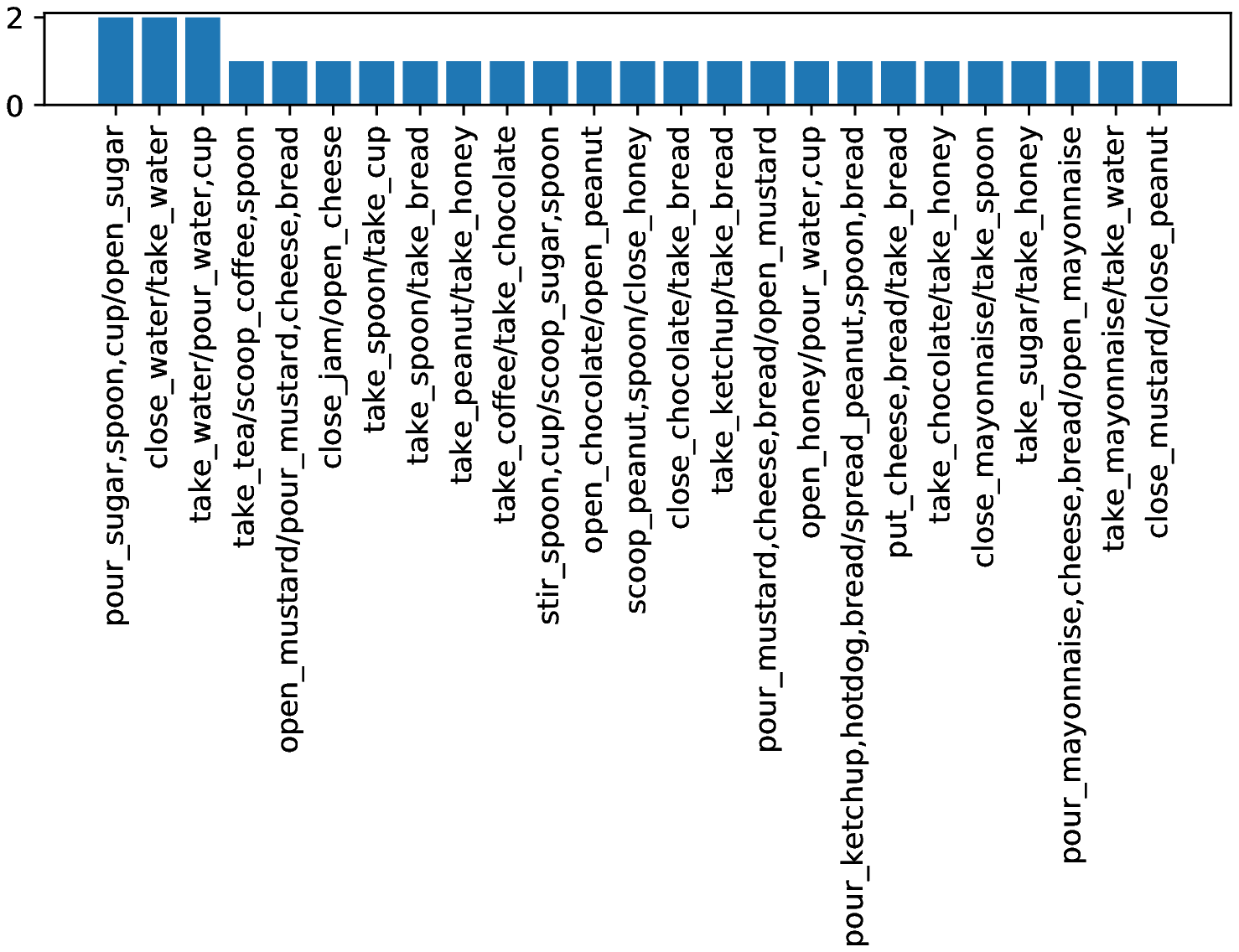}
			\caption{}
			\label{fig:base+poolingImp}
		\end{subfigure} \hspace{4mm}
		\begin{subfigure}[b]{0.45\textwidth}
			\includegraphics[scale=0.53]{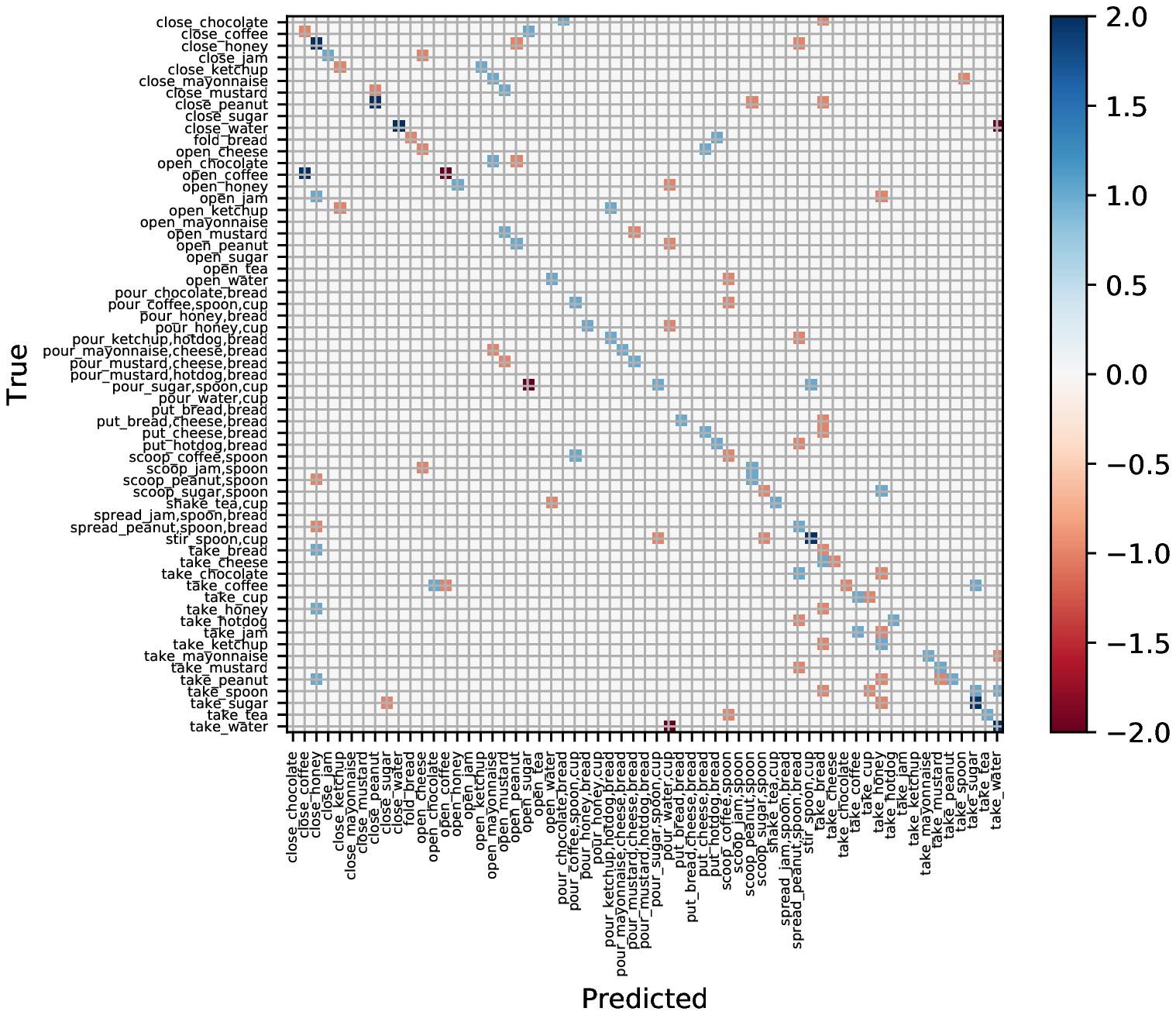}
			\label{fig:base+poolingConf}
			\caption{}
		\end{subfigure}
		\caption{Most improvement categories by adding both attention and output pooling to the baseline on GTEA 61 fixed split. X axis labels are in the format true label (baseline + pooling)/predicted label (baseline). Y axis shows the number of corrected samples for each class. (b) shows the difference of confusion matrices.}
		\label{fig:base+pooling}
	\end{figure*}

\begin{figure*}[t]
		\centering
		\begin{subfigure}[b]{0.45\textwidth}
			\includegraphics[scale=0.53]{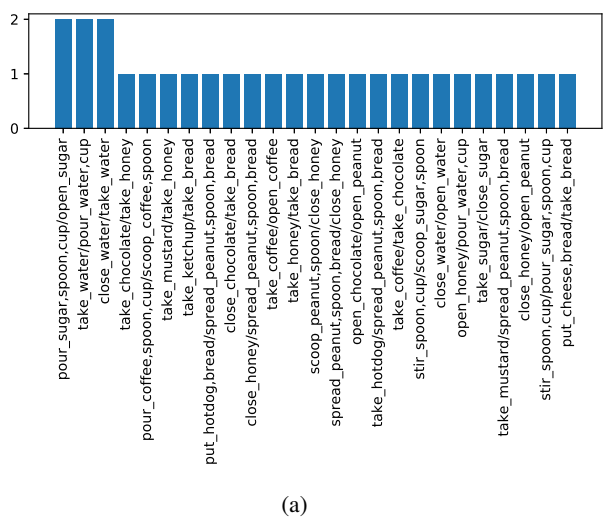}
			\caption{}
			\label{fig:CLSTM_LSTAImp}
		\end{subfigure} \hspace{4mm}
		\begin{subfigure}[b]{0.45\textwidth}
			\includegraphics[scale=0.53]{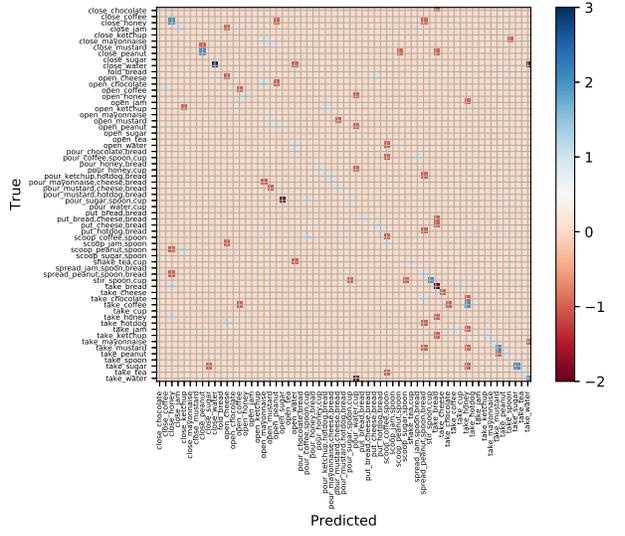}
			\label{fig:CLSTM_LSTAConf}
			\caption{}
		\end{subfigure}
		\caption{Most improvement categories by adding attention and output pooling with bias control (full LSTA model) to the baseline on GTEA 61 fixed split. X axis labels are in the format true label (LSTA)/predicted label (baseline). Y axis shows the number of corrected samples for each class. (b) shows the difference of confusion matrices.}
		\label{fig:CLSTM_LSTA}
	\end{figure*}

	Figs.~\ref{fig:base+outPooling}~-~\ref{fig:CLSTM_LSTA} show details of the classes which are improved by proposed LSTA variants over the baseline (ConvLSTM) and the difference of the confusion matrices. We show the top 25 improved classes in the comparison graphs and those with less number list all the improved classes. The difference of confusion matrices show the overall details of the classes which are improved. Ideally, the positive values should be in the diagonal and the negative values off-diagonal. Tab.~\ref{tab:action+obj} lists a breakdown of the recognition performance. For this, we compute the action recognition and object recognition performance of a network trained for activity recognition. There are some activity classes with multiple objects and these objects are combined to form a meta-object class for this analysis.

	Fig.~\ref{fig:base+outPooling} compares the baseline (ConvLSTM) with a network having baseline+output pooling, as explained in Sec.~4.2. It can be seen that adding output pooling to the ConvLSTM improves the network's capability in recognizing different actions with the same objects (\verb+take_water/pour_water,cup+ and \verb+close_water/take_water+). This confirms our hypothesis 
	that the output gating of LSTM affects memory tracking, replacing the output gating of LSTM with the proposed output pooling technique localizes the active memory component. This improves the tracking of relevant spatio-temporal patterns in the memory and consequently boosts recognition performance. A gain of $13.79\%$ is achieved for action recognition as shown in Tab.~\ref{tab:action+obj}.
	
	\begin{table*}[t!]\small
	\centering
	\begin{tabular}{|l|c|c|c|}
		\hline
		\multirow{2}{*}{Method} & \multicolumn{3}{c|}{Accuracy (\%)}\\ \cline{2-4}
		& Activity & Action & Object\\ 
		\hline \hline
        Baseline & 51.72 & 65.52 & 57.76\\
        \hline
        Baseline+output pooling & 62.07 & 79.31 ({\bf+13.79}) & 69.83 ({+12,07})\\ \hline
        Baseline+attention pooling & 66.38 & 78.45 (+12,93) & 74.14 ({\bf+16,38}) \\ \hline
        Baseline+pooling & 68.1 & 79.31 (+13.79) & 75.86 (+18,10)\\ \hline
        LSTA & 74.14 & 87.93  ({\bf+22.41}) & 79.31 (\bf+21,55) \\ \hline
	\end{tabular}
	\caption{Detailed ablation analysis on GTEA 61 fixed split. We compute the action and object recognition score by decomposing the action and objects from the predicted activity label.}
	\label{tab:action+obj}
\end{table*}

	In Fig.~\ref{fig:base+attPooling}, we can see that the network with the attention pooling described in Sec.~4.1 improves the categories with different actions and same objects as well as activity classes with multiple objects (\verb+stir_spoon,cup/pour_sugar,spoon,cup+; \verb+put_cheese,bread/take_bread+; \verb+pour_coffee,spoon,cup/scoop_coffee,spoon+, etc.). 
Attention helps the network to encode the features from the spatially relevant areas. This allows the network to keep a track of the active object regions and improves the performance. From Tab.~\ref{tab:action+obj}, a gain of $20.69\%$ is obtained for object recognition which gives further validation regarding the importance of attention.  

Adding both attention pooling and output pooling further improves the network's capability in distinguishing between different actions with same objects and same actions with different objects. This is visible in Fig.~\ref{fig:base+pooling} and also from the $13.72\%$ and $18.1\%$ performance gain obtained for action and object recognition, respectively. 

Incorporating bias control, introduced in Sec.~4.2, to the output pooling results in the proposed method, \acs{lsta}, which further improves the capacity of the network in recognizing activities (Fig.~\ref{fig:CLSTM_LSTA}). This further verifies the hypothesis in Sec.~4.2 that bias control increases the active memory localization of the network. This is also evident from Tab.~\ref{tab:action+obj} where an increase of $22.41\%$ is obtained for action recognition.

It is worth noting that output pooling boosts action recognition performance more (+13.79\% action vs +12,07\% object) while with attention pooling the object recognition performance receives a higher gain (+12,93\% vs +16,38\%). Coupling attention and output pooling through bias control finally boosts performance by a significant margin on both (+22.41\% vs +21,55\%).
This provides further evidence that the two contributions are complementary and reflects the intuitions behind the design choices of LSTA, making the improvements explainable and the benefits of each of the contributions transparently confirmed by this analysis.


	\section{Comparative Analysis}
	
\begin{figure*}[t]
		\centering
		\begin{subfigure}[b]{0.45\textwidth}
			\includegraphics[scale=0.53]{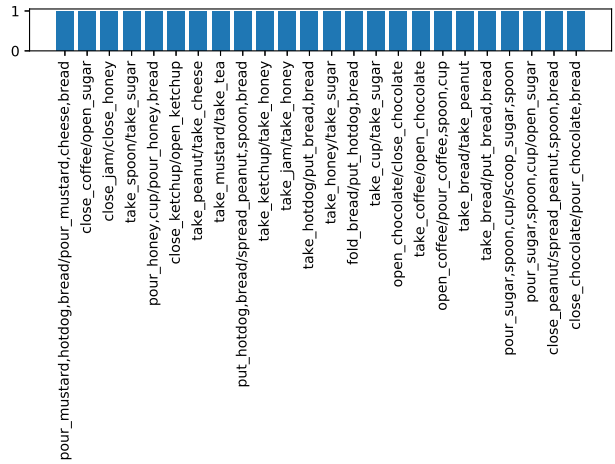}
			\caption{}
			\label{fig:egoRNN_LSTAImp}
		\end{subfigure} \hspace{4mm}
		\begin{subfigure}[b]{0.45\textwidth}
			\includegraphics[scale=0.53]{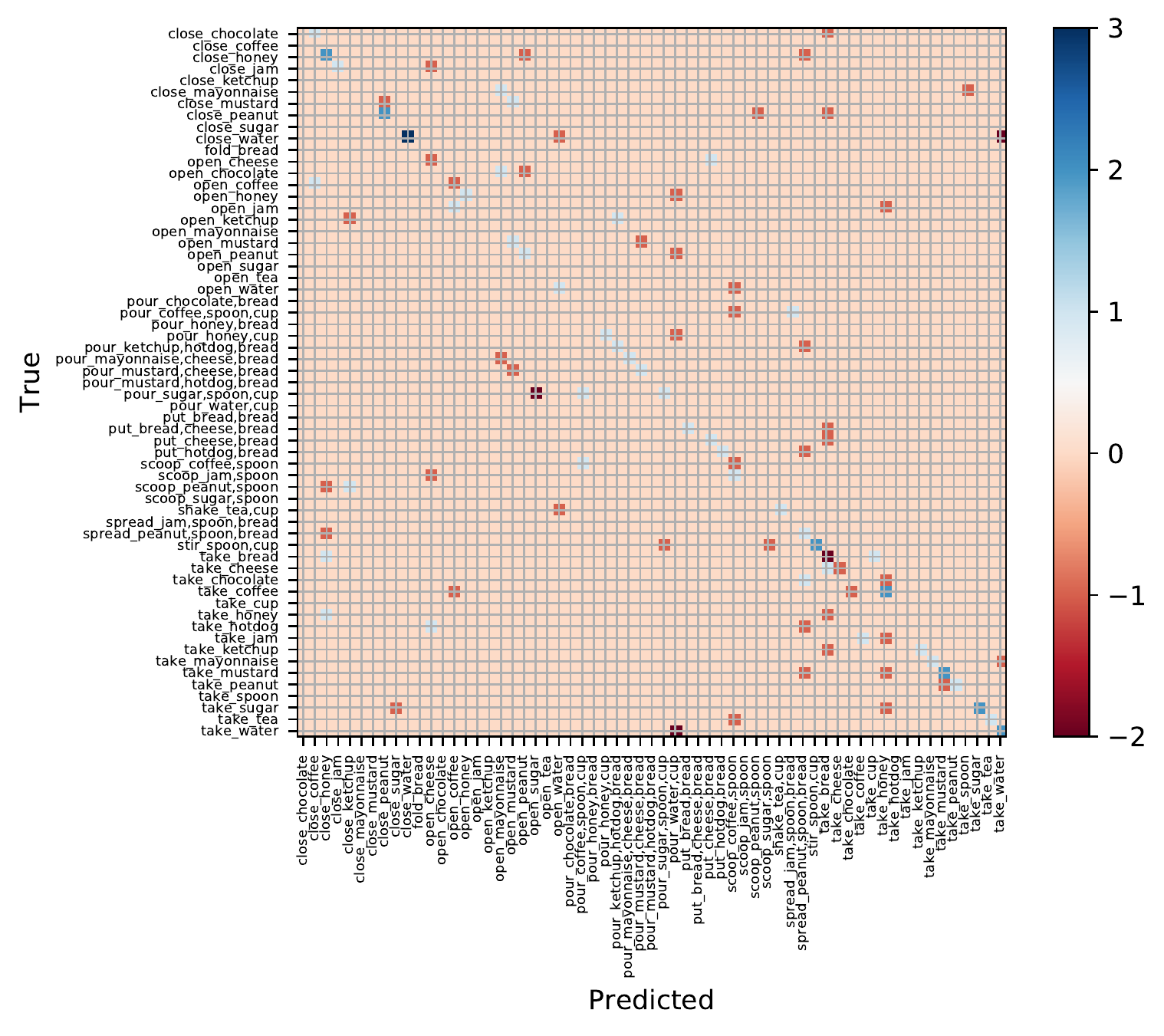}
			\label{fig:egoRNN_LSTAConf}
			\caption{}
		\end{subfigure}
		\caption{(a) Most improvement categories by LSTA over ego-rnn on GTEA 61 fixed split. X axis labels are in the format true label (LSTA)/predicted label (ego-rnn). Y axis shows the number of corrected samples for each class. (b) shows the difference of confusion matrices.}
		\label{fig:egoRNN_LSTA}
	\end{figure*}

\begin{figure*}[t]
		\centering
		\begin{subfigure}[b]{0.45\textwidth}
			\includegraphics[scale=0.53]{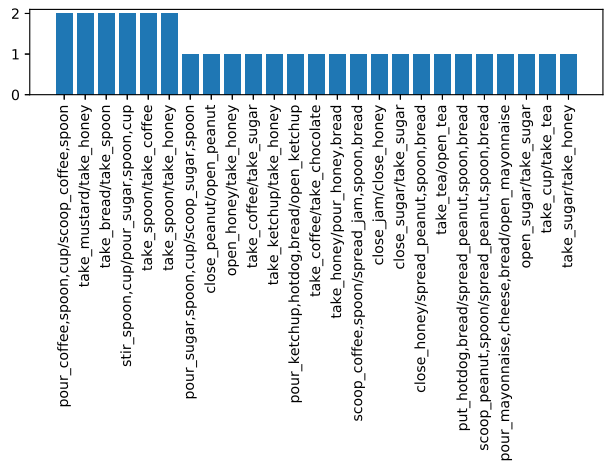}
			\caption{}
			\label{fig:eleGAttLSTM_LSTAImp}
		\end{subfigure} \hspace{4mm}
		\begin{subfigure}[b]{0.45\textwidth}
			\includegraphics[scale=0.53]{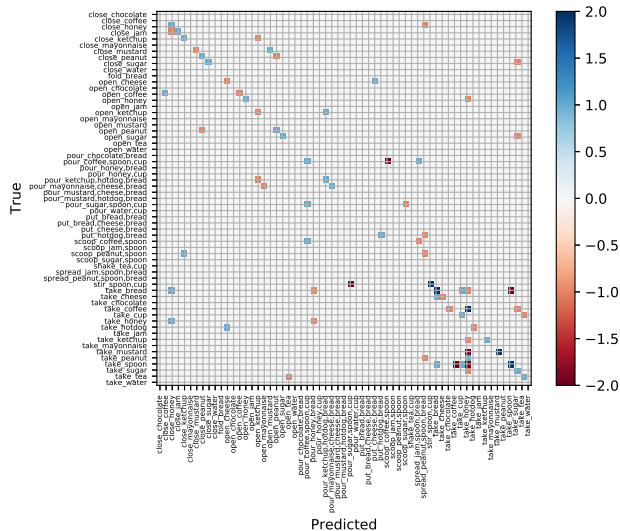}
			\label{fig:eleGAttLSTM_LSTAConf}
			\caption{}
		\end{subfigure}
		\caption{(a) Most improvement categories by LSTA over eleGAtt-LSTM on GTEA 61 fixed split. X axis labels are in the format true label (LSTA)/predicted label (eleGAtt-LSTM). Y axis shows the number of corrected samples for each class. (b) shows the difference of confusion matrices.}
		\label{fig:eleGAttLSTM_LSTA}
	\end{figure*}

\begin{figure*}[t]
		\centering
		\begin{subfigure}[b]{0.45\textwidth}
			\includegraphics[scale=0.53]{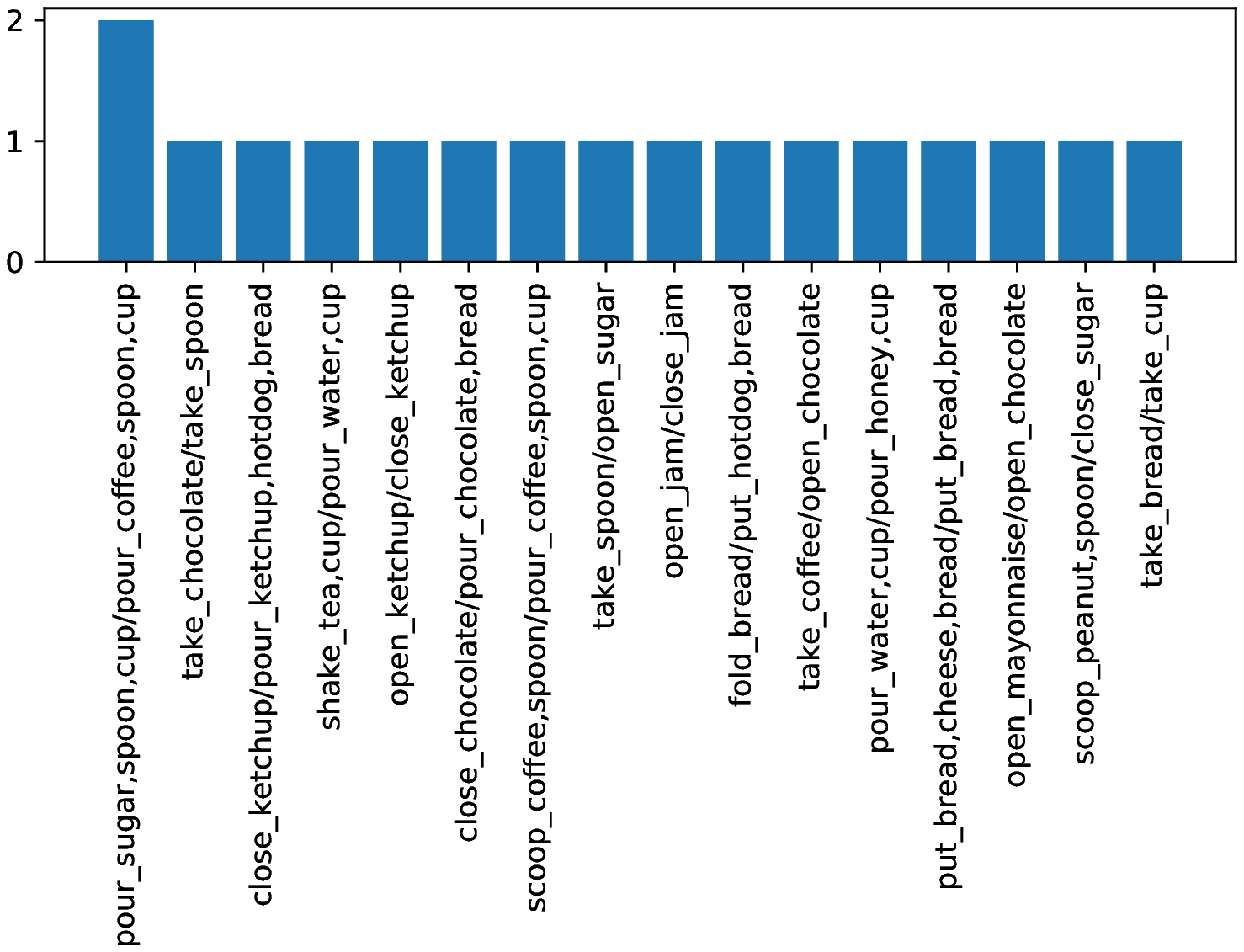}
			\caption{}
			\label{fig:twoStream_crossmodal_impImp}
		\end{subfigure} \hspace{4mm}
		\begin{subfigure}[b]{0.45\textwidth}
			\includegraphics[scale=0.53]{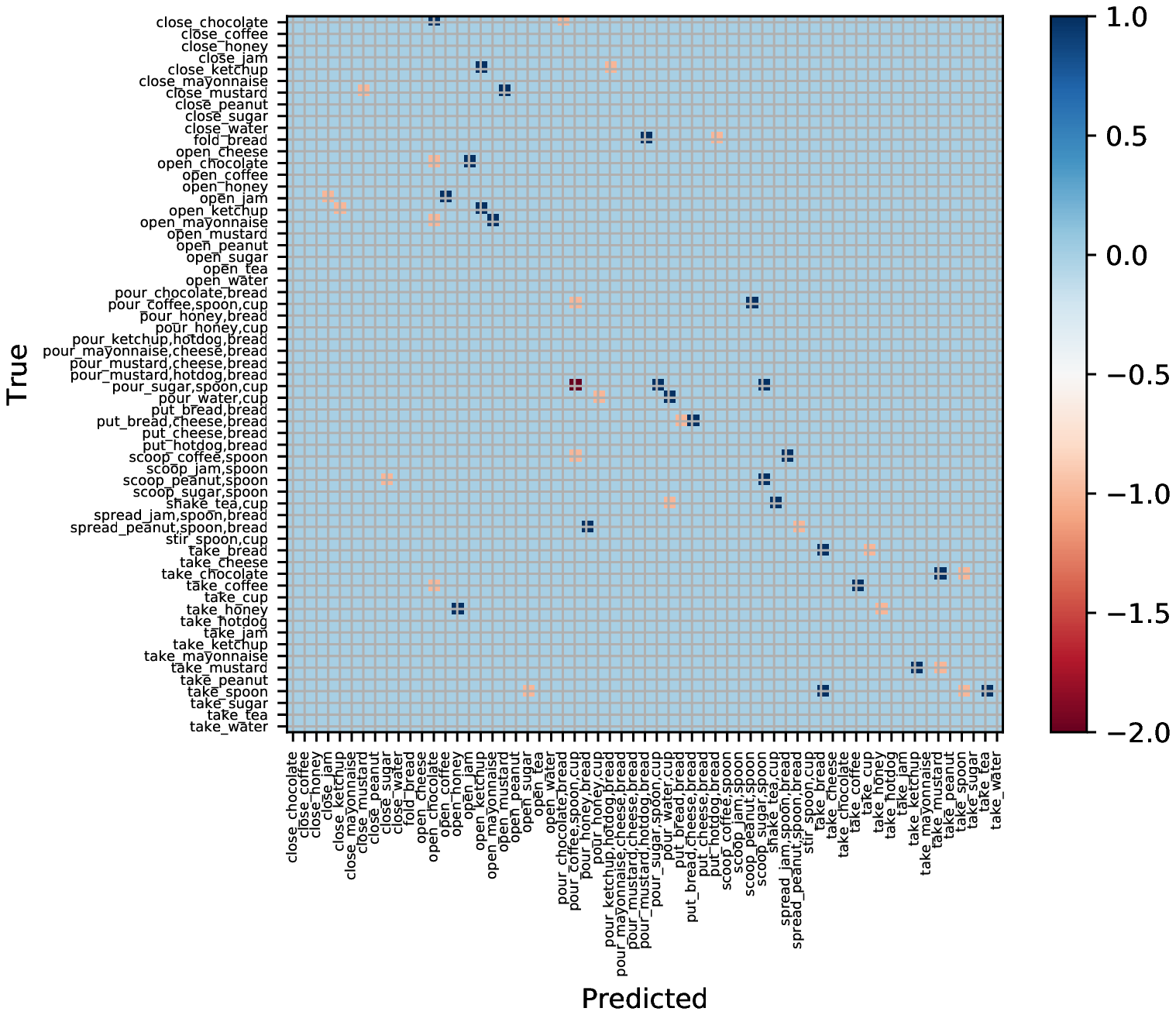}
			\label{fig:twoStream_crossmodal_impConf}
			\caption{}
		\end{subfigure}
		\caption{(a) Most improvement categories by two stream cross-modal fusion over two stream on GTEA 61 fixed split. X axis labels are in the format true label (two stream cross-modal fusion)/predicted label (two stream late fusion). Y axis shows the number of corrected samples for each class. (b) shows the difference of confusion matrices.}
		\label{fig:twoStream_crossmodal}
	\end{figure*}

	Figs.~\ref{fig:egoRNN_LSTA}~-~\ref{fig:twoStream_crossmodal} compares our method with state-of-the-art alternatives discussed in Sec.~2.3, ego-rnn~\cite{sudhakaran2018attention} and eleGatt~\cite{attention_eccv18}. Compared to ego-rnn, LSTA is capable of identifying activities involving multiple objects (\verb+pour_mustard,hotdog,bread/pour_mustard,+ \verb+cheese,bread+; \verb+pour_honey,cup/pour_honey,+ \verb+bread+; \verb+put_hotdog,bread/spread_peanut,+ \verb+spoon,bread+, etc.). This may be attributed to the attention mechanism with memory for tracking previously attended regions, helping the network attending to the same objects in subsequent frames. From Fig.~\ref{fig:eleGAttLSTM_LSTA} it can be seen that eleGAtt-LSTM fails to identify the objects correctly (\verb+take_mustard/take_honey+; \verb+take_bread/take_spoon+; \verb+take_spoon/take_honey+, etc.). This shows the attention map generated by LSTA selects more relevant regions compared to eleGAtt-LSTM .

	\section{Confusion Matrix}
	
\begin{figure*}[t]
		\centering
		\includegraphics[scale=1.1]{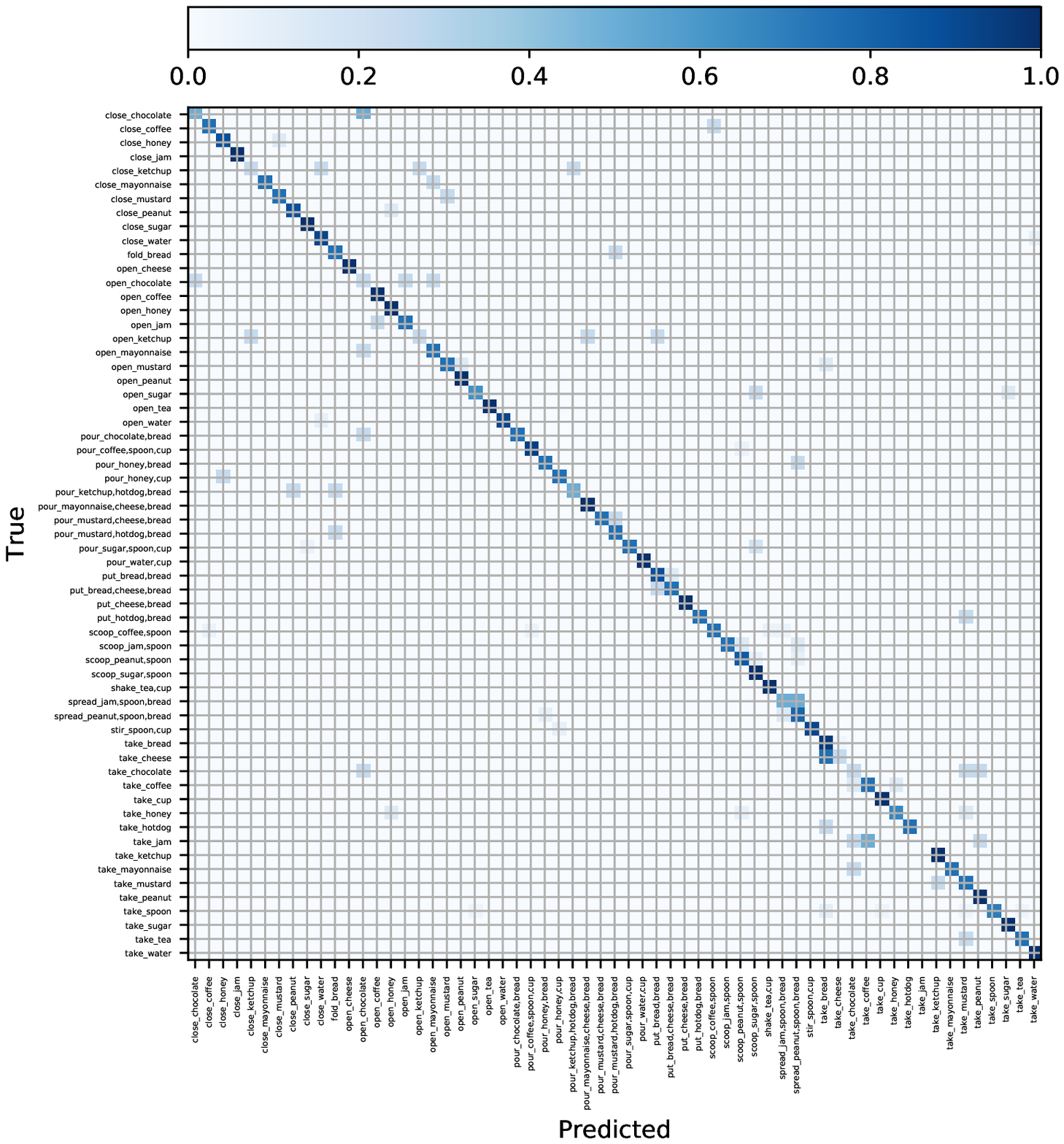}
		\caption{Confusion matrix of GTEA 61 averaged across the four train/test splits.}
		\label{fig:twoStream_crossmodal_gtea61_conf}
	\end{figure*}

\begin{figure*}[t]
		\centering
		\includegraphics[scale=1.1]{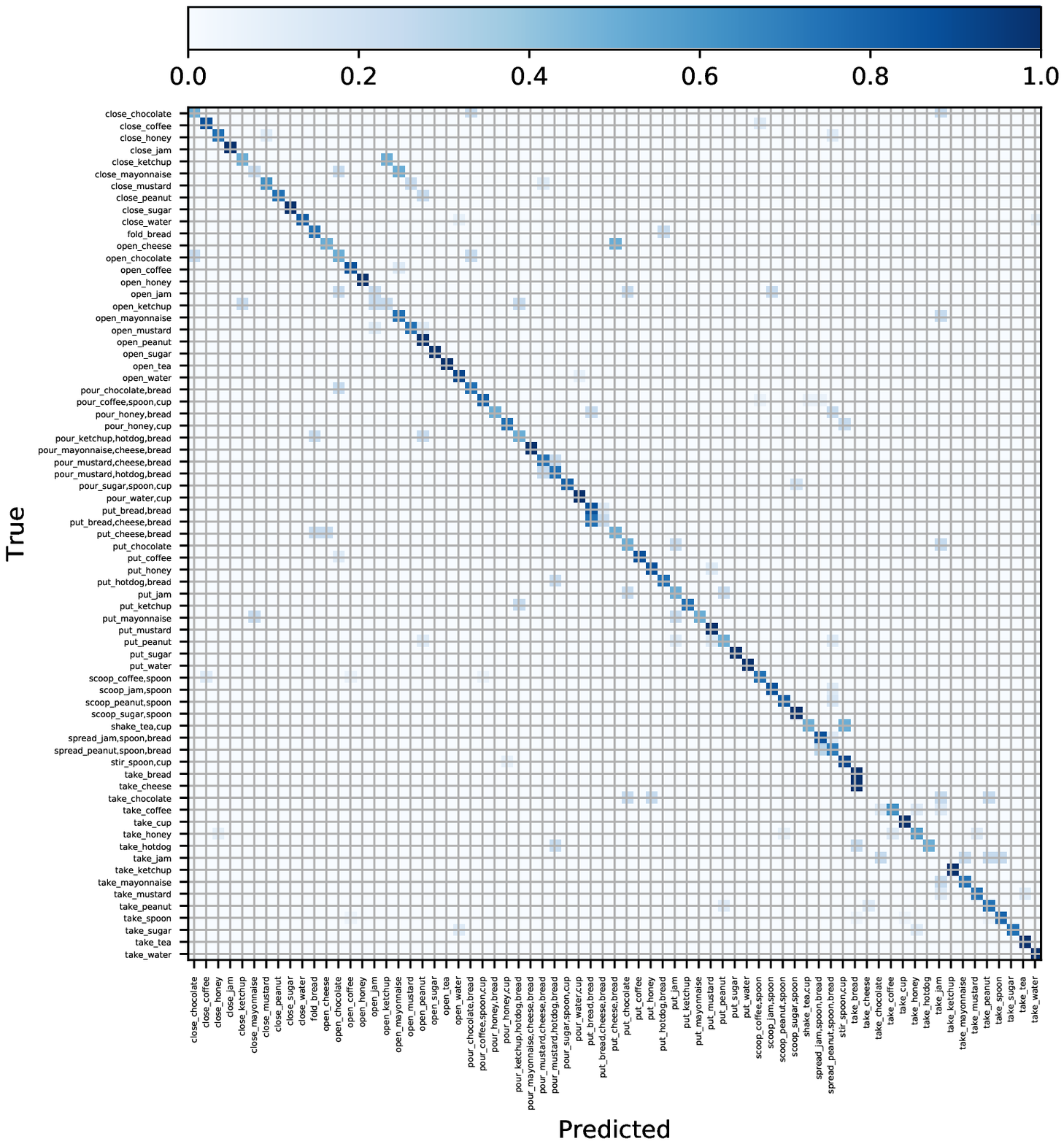}
		\caption{Confusion matrix of GTEA 71 averaged across the four train/test splits.}
		\label{fig:twoStream_crossmodal_gtea71_conf}
	\end{figure*}

\begin{figure*}[t]
		\centering
		\includegraphics[scale=1.05]{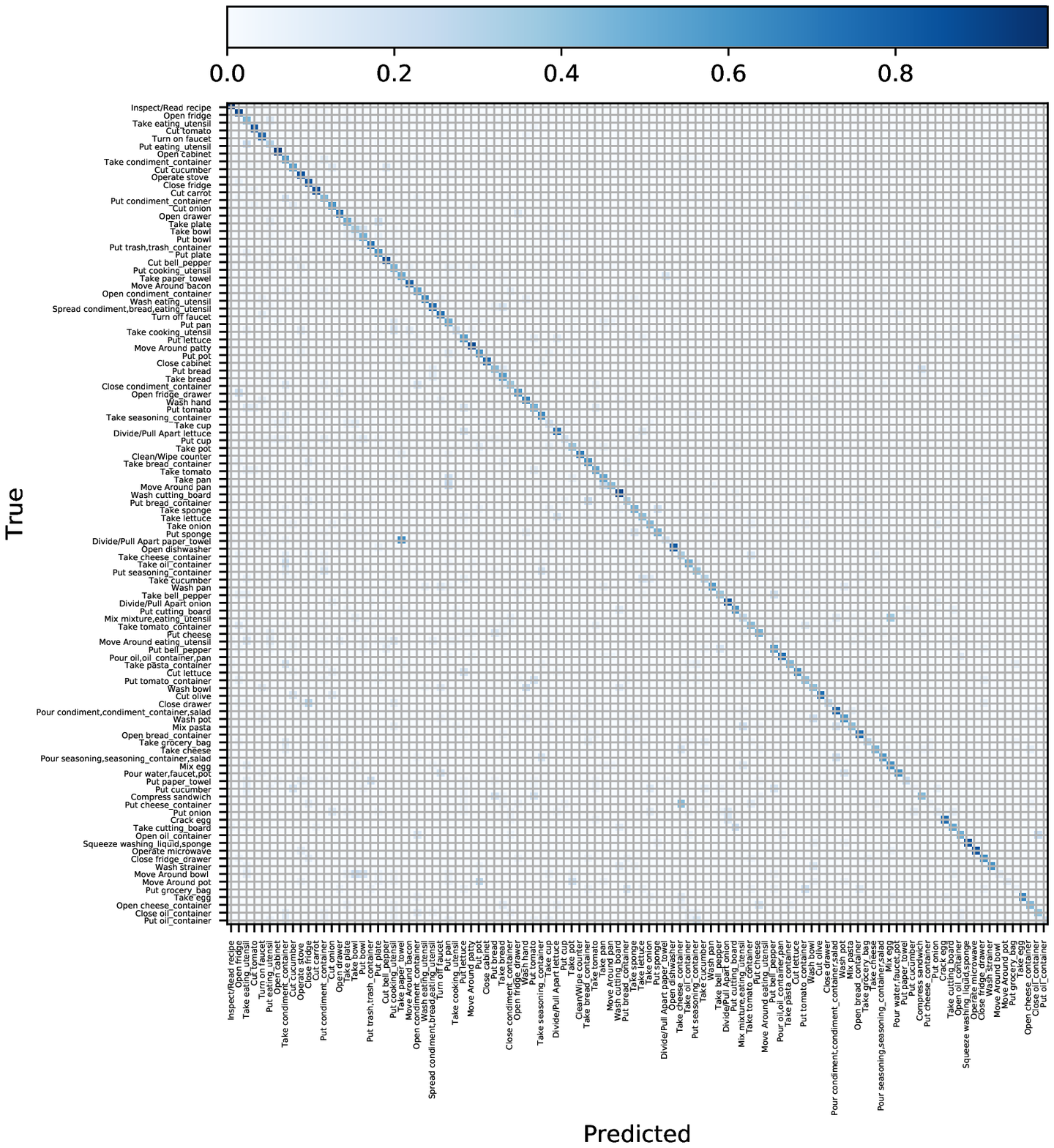}
		\caption{Confusion matrix of EGTEA Gaze+ averaged across the three train/test splits.}
		\label{fig:twoStream_crossmodal_egtea_conf}
	\end{figure*}

	Figs.~\ref{fig:twoStream_crossmodal_gtea61_conf}~-~\ref{fig:twoStream_crossmodal_egtea_conf} show the confusion matrix of the LSTA (two stream cross-modal fusion) for all the datasets explained in Sec.~6.1 of the manuscript. We average the confusion matrices of each of the available train/test splits to generate a single confusion matrix representing the dataset under consideration.

	\section{EPIC-KITCHENS}
	We compare the recognition accuracies obtained for EPIC-KITCHENS dataset with the currently available baselines \cite{Damen_2018_ECCV} in Tab.~\ref{tab:epic_kitchens}. As explained in Sec.~6.6 in the paper, we train the network for predicting \verb+verb+ and \verb+noun+ and \verb+activity+ classes. Our two stream cross-modal fusion model obtains an activity recognition performance of $30.33\%$ and $16.63\%$ on S1 and S2 settings as opposed to the $20.54\%$ and $10.89\%$ obtained by TSN strongest baseline (two stream). It is also worth noting that our model is strong on predicting \verb+verb+ ($+11.32\%$ points on S1 setting over strongest baseline). This indicates LSTA accurately performs encoding of sequences, indeed \verb+verb+ in this context is typically describing actions that develop into an activity over time, and this is learned effectively with LSTA just using video-level supervision.
\begin{table*}[t]\small
	\centering
	\begin{tabular}{c|l|c|c|c|c|c|c|c|c|c|c|c|c}
		\hline
		& Method & \multicolumn{3}{c|}{Top-1 Accuracy (\%)} & \multicolumn{3}{c|}{Top-5 Accuracy (\%)} & \multicolumn{3}{c|}{Precision (\%)} & \multicolumn{3}{c}{Recall (\%)} \\
		\cline{3-14}
		& & \parbox{0.7cm}{Verb} & \parbox{0.7cm}{Noun} & \parbox{0.8cm}{Action} & \parbox{0.7cm}{Verb} & \parbox{0.7cm}{Noun} & \parbox{0.8cm}{Action} & \parbox{0.7cm}{Verb} & \parbox{0.7cm}{Noun} & \parbox{0.8cm}{Action} & \parbox{0.7cm}{Verb} & \parbox{0.7cm}{Noun} & \parbox{0.8cm}{Action}\\
		\hline \hline
		\multirow{4}{*}{\rotatebox[origin=t]{90}{S1}} & 2SCNN (RGB) & 40.44 & 30.46 & 13.67 & 83.04 & 57.05 & 33.25 & 34.74 & 28.23 & 6.66 & 15.90 & 23.23 & 5.47\\ \cline{2-14}
		& 2SCNN (two stream) & 42.16 & 29.14 & 13.23 & 80.58 & 53.70 & 30.36 & 29.39 & 30.73 & 5.92 & 14.83 & 21.10 & 4.93\\ \cline{2-14}
		& TSN (RGB) & 45.68 & 36.80 & 19.86 & 85.56 & 64.19 & 41.89 & \textbf{61.64} & 34.32 & 11.02 & 23.81 & 31.62 & 9.76\\ \cline{2-14}
		& TSN (two stream) & 48.23 & 36.71 & 20.54 & 84.09 & 62.32 & 39.79 & 47.26 & 35.42 & 11.57 & 22.33 & 30.53 & 9.78\\ \cline{2-14}
		& \textbf{LSTA (RGB)} & 58.25 & \textbf{38.93} & 30.16 & \textbf{86.57} & \textbf{62.96} & \textbf{50.16} & 44.09 & \textbf{36.30} & \textbf{16.54} & 37.32 & \textbf{36.52} & \textbf{19.00}\\ \cline{2-14}
		& \textbf{LSTA (two stream)} & \textbf{59.55} & 38.35 & \textbf{30.33} & 85.77 & 61.49 & 49.97 & 42.72 & 36.19 & 14.46 & \textbf{38.12} & 36.19 & 17.76\\
	 \hline
        \multirow{4}{*}{\rotatebox[origin=t]{90}{S2}} & 2SCNN (RGB) & 34.89 & 21.82 & 10.11 & 74.56 & 45.34 & 25.33 & 19.48 & 14.67 & 5.32 & 11.22 & 17.24 & 6.34\\ \cline{2-14}
		& 2SCNN (two stream) & 36.16 & 18.03 & 7.31 & 71.97 & 38.41 & 19.49 & 18.11 & 15.31 & 3.19 & 10.52 & 12.55 & 3.00\\ \cline{2-14}
		& TSN (RGB) & 34.89 & 21.82 & 10.11 & 74.56 & 45.34 & 25.33 & 19.48 & 14.67 & 5.32 & 11.22 & 17.24 & 6.34\\ \cline{2-14}
 		& TSN (two stream)& 39.4 & 22.7 & 10.89 & 74.29 & \textbf{45.72} & 25.26 & 22.54 & 15.33 & 6.21 & 13.06 & 17.52 & 6.49\\ \cline{2-14}
 		& \textbf{LSTA (RGB)} & 45.51 & \textbf{23.46} & 15.88 & 75.25 & 43.16 & 30.01 & 26.19 & 17.58 & 8.44 & 20.80 & \textbf{19.67} & 11.29\\ \cline{2-14}
 		& \textbf{LSTA (two stream)} & \textbf{47.32} & 22.16 & \textbf{16.63} & \textbf{77.02} & 43.15 & \textbf{30.93} & \textbf{31.57} & \textbf{17.91} & \textbf{8.97} & \textbf{26.17} & 17.80 & \textbf{11.92}\\ \hline \cline{2-14}
	\end{tabular}
	\caption{Comparison of recognition accuracies with state-of-the-art in EPIC-KITCHENS dataset.}
	\label{tab:epic_kitchens}
\end{table*}
	

\section{Attention Map Visualization}

\begin{figure*}[t]
	\centering      
	\raisebox{.5in}{\rotatebox[origin=t]{90}{Input}}
	\begin{subfigure}[b]{0.95\textwidth}
		\includegraphics[scale=0.4]{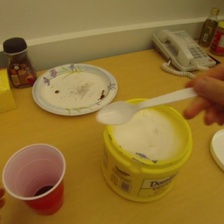}
		\includegraphics[scale=0.4]{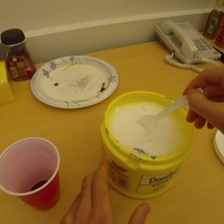}
		\includegraphics[scale=0.4]{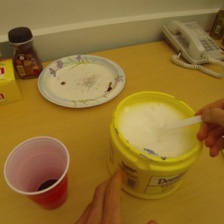}
		\includegraphics[scale=0.4]{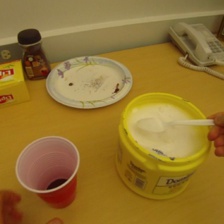}
		\includegraphics[scale=0.4]{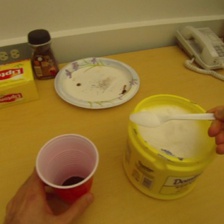}
	\end{subfigure}\\ \vskip 2mm
	\ \raisebox{.5in}{\rotatebox[origin=t]{90}{ego-rnn}}
	\begin{subfigure}[b]{0.95\textwidth}
		\includegraphics[scale=0.4]{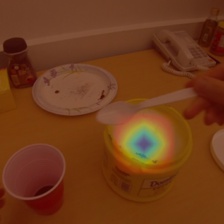}
		\includegraphics[scale=0.4]{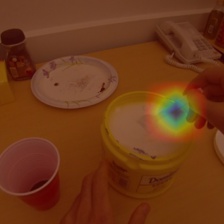}
		\includegraphics[scale=0.4]{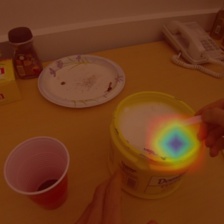}
		\includegraphics[scale=0.4]{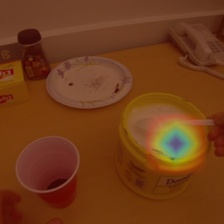}
		\includegraphics[scale=0.4]{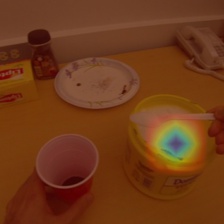}
	\end{subfigure}\\ \vskip 2mm
	\ \raisebox{.5in}{\rotatebox[origin=t]{90}{LSTA}}
	\begin{subfigure}[b]{0.95\textwidth}
		\includegraphics[scale=0.4]{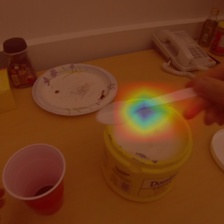}
		\includegraphics[scale=0.4]{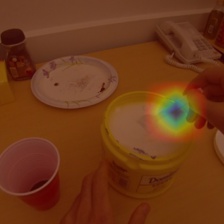}
		\includegraphics[scale=0.4]{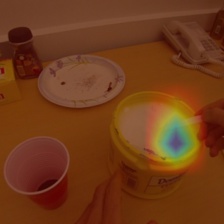}
		\includegraphics[scale=0.4]{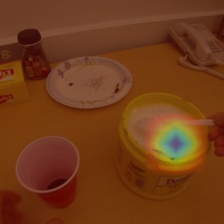}
		\includegraphics[scale=0.4]{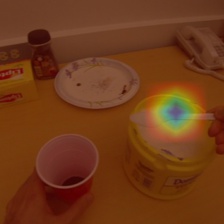}
	\end{subfigure}\\ \vskip 2mm	       
	\ \raisebox{.5in}{\rotatebox[origin=t]{90}{Flow}}
	\begin{subfigure}[b]{0.95\textwidth}
		\includegraphics[scale=0.4]{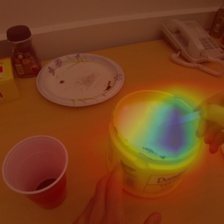}
		\includegraphics[scale=0.4]{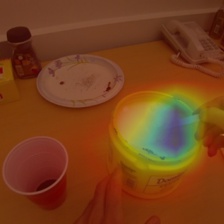}
		\includegraphics[scale=0.4]{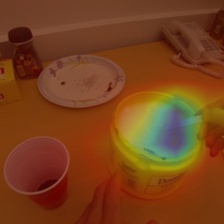}
		\includegraphics[scale=0.4]{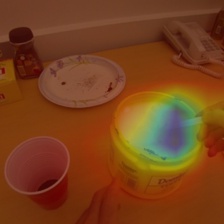}
		\includegraphics[scale=0.4]{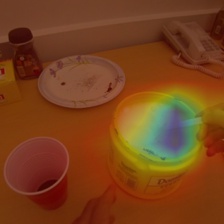}
	\end{subfigure}\\ \vskip 2mm			 
	\caption{Attention maps generated by ego-rnn (second row) and LSTA (third) for scoop\_sugar,spoon video sequence. We show the 5 frames that are uniformly sampled from the 25 frames used as input to the corresponding networks. Fourth row shows the attention map generated by the motion stream. For flow, we visualize the attention map on the five frames corresponding to the optical flow stack given as input.}
	\label{fig:att_map_scoop_sugar}
\end{figure*}

\begin{figure*}[t]
	\centering      
	\raisebox{.5in}{\rotatebox[origin=t]{90}{Input}}
	\begin{subfigure}[b]{0.95\textwidth}
		\includegraphics[scale=0.4]{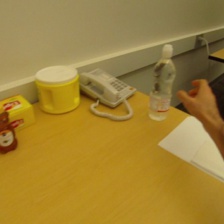}
		\includegraphics[scale=0.4]{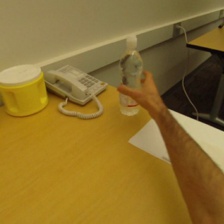}
		\includegraphics[scale=0.4]{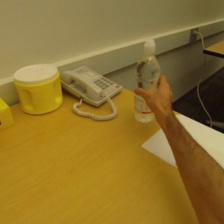}
		\includegraphics[scale=0.4]{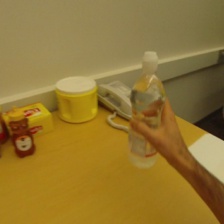}
		\includegraphics[scale=0.4]{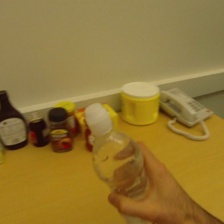}
	\end{subfigure}\\ \vskip 2mm
	\ \raisebox{.5in}{\rotatebox[origin=t]{90}{ego-rnn}}
	\begin{subfigure}[b]{0.95\textwidth}
		\includegraphics[scale=0.4]{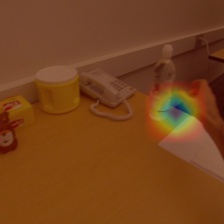}
		\includegraphics[scale=0.4]{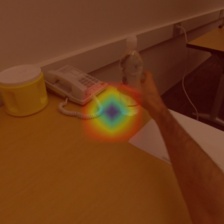}
		\includegraphics[scale=0.4]{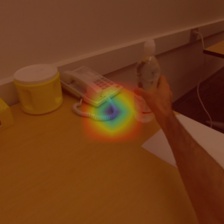}
		\includegraphics[scale=0.4]{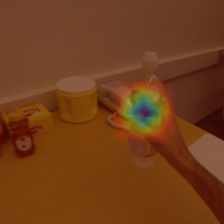}
		\includegraphics[scale=0.4]{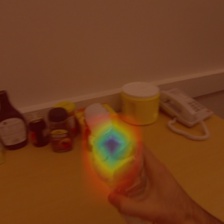}
	\end{subfigure}\\ \vskip 2mm
	\ \raisebox{.5in}{\rotatebox[origin=t]{90}{LSTA}}
	\begin{subfigure}[b]{0.95\textwidth}
		\includegraphics[scale=0.4]{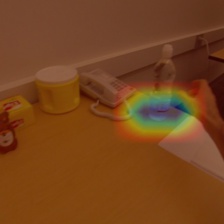}
		\includegraphics[scale=0.4]{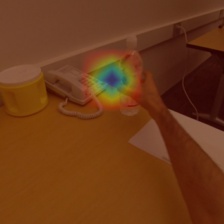}
		\includegraphics[scale=0.4]{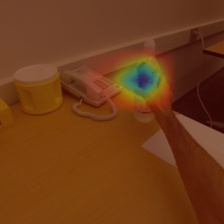}
		\includegraphics[scale=0.4]{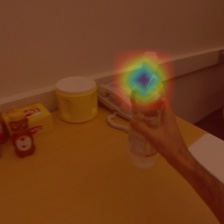}
		\includegraphics[scale=0.4]{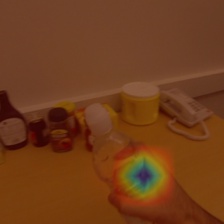}
	\end{subfigure}\\ \vskip 2mm	       
	\ \raisebox{.5in}{\rotatebox[origin=t]{90}{Flow}}
	\begin{subfigure}[b]{0.95\textwidth}
		\includegraphics[scale=0.4]{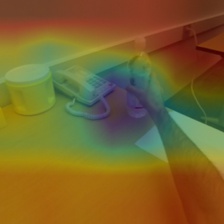}
		\includegraphics[scale=0.4]{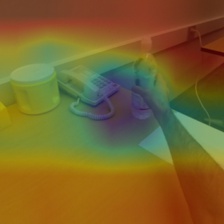}
		\includegraphics[scale=0.4]{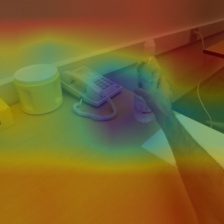}
		\includegraphics[scale=0.4]{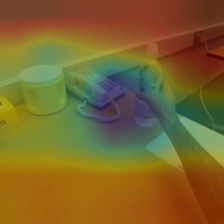}
		\includegraphics[scale=0.4]{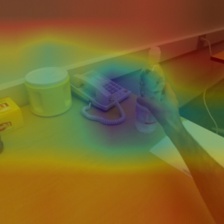}
	\end{subfigure}\\ \vskip 2mm			 
	\caption{Attention maps generated by ego-rnn (second row) and LSTA (third) for take\_water video sequence. We show the 5 frames that are uniformly sampled from the 25 frames used as input to the corresponding networks. Fourth row shows the attention map generated by the motion stream. For flow, we visualize the attention map on the five frames corresponding to the optical flow stack given as input.}
	\label{fig:att_map_take_water}
\end{figure*}

\begin{figure*}[t]
	\centering      
	\raisebox{.5in}{\rotatebox[origin=t]{90}{Input}}
	\begin{subfigure}[b]{0.95\textwidth}
		\includegraphics[scale=0.4]{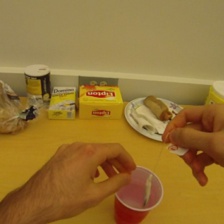}
		\includegraphics[scale=0.4]{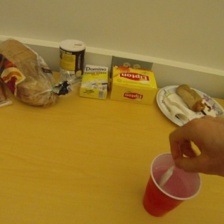}
		\includegraphics[scale=0.4]{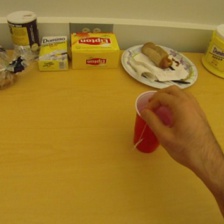}
		\includegraphics[scale=0.4]{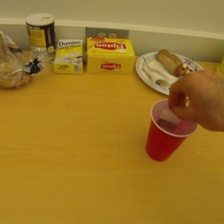}
		\includegraphics[scale=0.4]{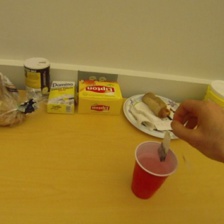}
	\end{subfigure}\\ \vskip 2mm
	\ \raisebox{.5in}{\rotatebox[origin=t]{90}{ego-rnn}}
	\begin{subfigure}[b]{0.95\textwidth}
		\includegraphics[scale=0.4]{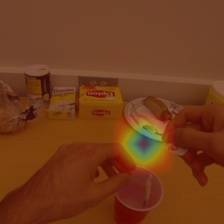}
		\includegraphics[scale=0.4]{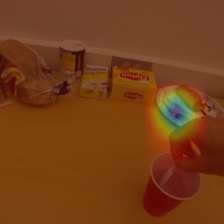}
		\includegraphics[scale=0.4]{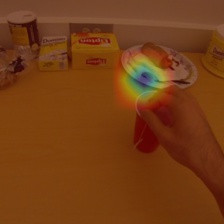}
		\includegraphics[scale=0.4]{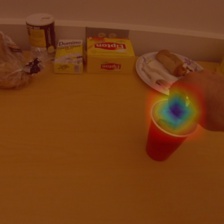}
		\includegraphics[scale=0.4]{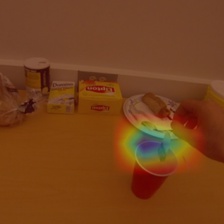}
	\end{subfigure}\\ \vskip 2mm
	\ \raisebox{.5in}{\rotatebox[origin=t]{90}{LSTA}}
	\begin{subfigure}[b]{0.95\textwidth}
		\includegraphics[scale=0.4]{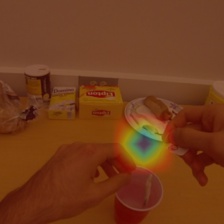}
		\includegraphics[scale=0.4]{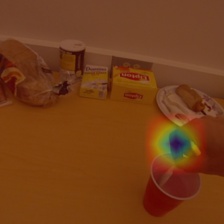}
		\includegraphics[scale=0.4]{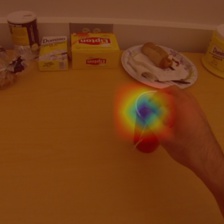}
		\includegraphics[scale=0.4]{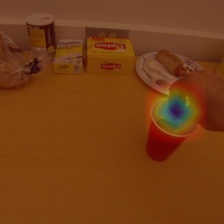}
		\includegraphics[scale=0.4]{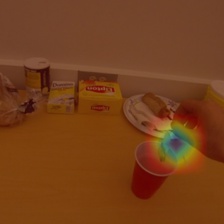}
	\end{subfigure}\\ \vskip 2mm	       
	\ \raisebox{.5in}{\rotatebox[origin=t]{90}{Flow}}
	\begin{subfigure}[b]{0.95\textwidth}
		\includegraphics[scale=0.4]{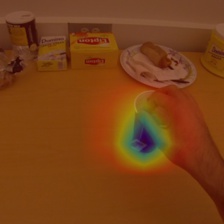}
		\includegraphics[scale=0.4]{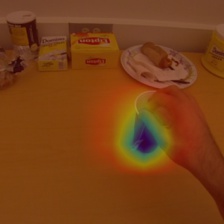}
		\includegraphics[scale=0.4]{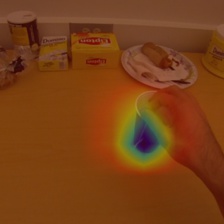}
		\includegraphics[scale=0.4]{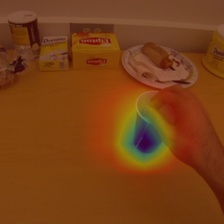}
		\includegraphics[scale=0.4]{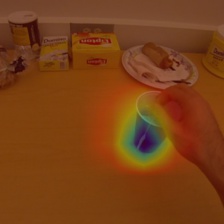}
	\end{subfigure}\\ \vskip 2mm			 
	\caption{Attention maps generated by ego-rnn (second row) and LSTA (third) for shake\_tea,cup video sequence. We show the 5 frames that are uniformly sampled from the 25 frames used as input to the corresponding networks. Fourth row shows the attention map generated by the motion stream. For flow, we visualize the attention map on the five frames corresponding to the optical flow stack given as input.}
	\label{fig:att_map_shake_tea}
\end{figure*}

\begin{figure*}[t]
	\centering      
	\raisebox{.5in}{\rotatebox[origin=t]{90}{Input}}
	\begin{subfigure}[b]{0.95\textwidth}
		\includegraphics[scale=0.4]{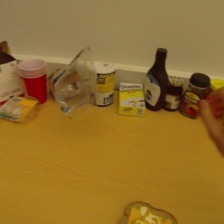}
		\includegraphics[scale=0.4]{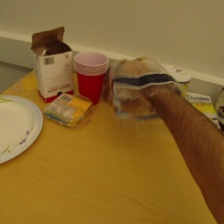}
		\includegraphics[scale=0.4]{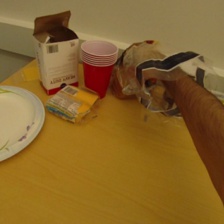}
		\includegraphics[scale=0.4]{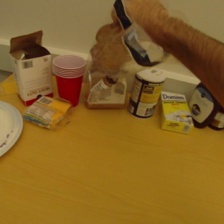}
		\includegraphics[scale=0.4]{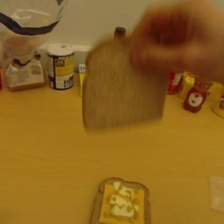}
	\end{subfigure}\\ \vskip 2mm
	\ \raisebox{.5in}{\rotatebox[origin=t]{90}{ego-rnn}}
	\begin{subfigure}[b]{0.95\textwidth}
		\includegraphics[scale=0.4]{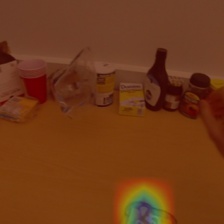}
		\includegraphics[scale=0.4]{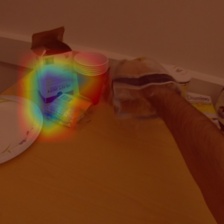}
		\includegraphics[scale=0.4]{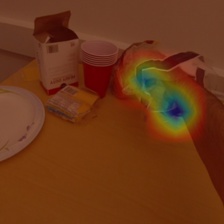}
		\includegraphics[scale=0.4]{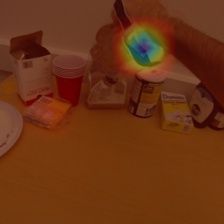}
		\includegraphics[scale=0.4]{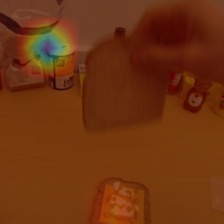}
	\end{subfigure}\\ \vskip 2mm
	\ \raisebox{.5in}{\rotatebox[origin=t]{90}{LSTA}}
	\begin{subfigure}[b]{0.95\textwidth}
		\includegraphics[scale=0.4]{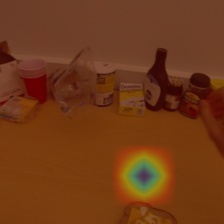}
		\includegraphics[scale=0.4]{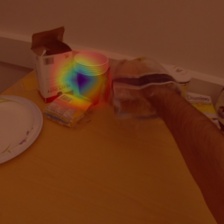}
		\includegraphics[scale=0.4]{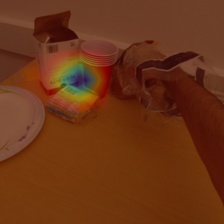}
		\includegraphics[scale=0.4]{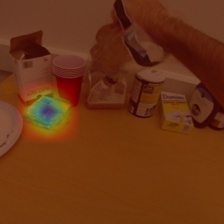}
		\includegraphics[scale=0.4]{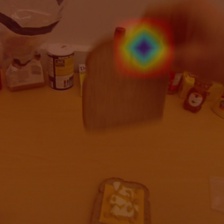}
	\end{subfigure}\\ \vskip 2mm	       
	\ \raisebox{.5in}{\rotatebox[origin=t]{90}{Flow}}
	\begin{subfigure}[b]{0.95\textwidth}
		\includegraphics[scale=0.4]{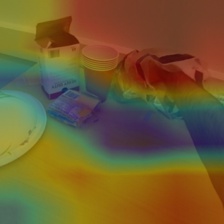}
		\includegraphics[scale=0.4]{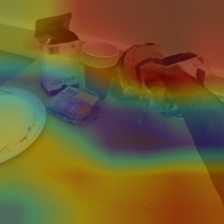}
		\includegraphics[scale=0.4]{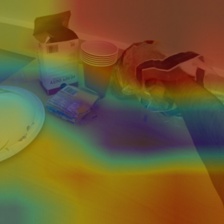}
		\includegraphics[scale=0.4]{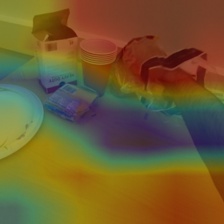}
		\includegraphics[scale=0.4]{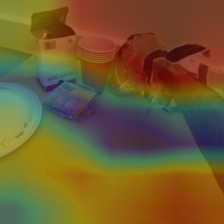}
	\end{subfigure}\\ \vskip 2mm			 
	\caption{Attention maps generated by ego-rnn (second row) and LSTA (third) for take\_bread video sequence. We show the 5 frames that are uniformly sampled from the 25 frames used as input to the corresponding networks. Fourth row shows the attention map generated by the motion stream. For flow, we visualize the attention map on the five frames corresponding to the optical flow stack given as input.}
	\label{fig:att_map_take_bread}
\end{figure*}

\begin{figure*}[t]
	\centering      
	\raisebox{.5in}{\rotatebox[origin=t]{90}{Input}}
	\begin{subfigure}[b]{0.95\textwidth}
		\includegraphics[scale=0.4]{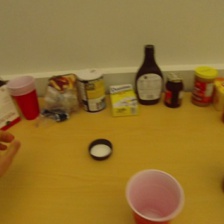}
		\includegraphics[scale=0.4]{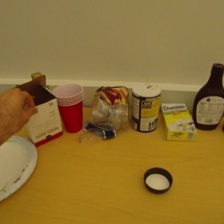}
		\includegraphics[scale=0.4]{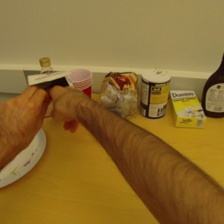}
		\includegraphics[scale=0.4]{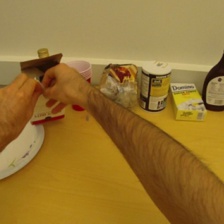}
		\includegraphics[scale=0.4]{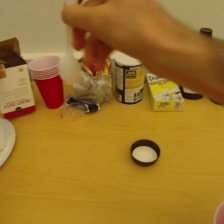}
	\end{subfigure}\\ \vskip 2mm
	\ \raisebox{.5in}{\rotatebox[origin=t]{90}{ego-rnn}}
	\begin{subfigure}[b]{0.95\textwidth}
		\includegraphics[scale=0.4]{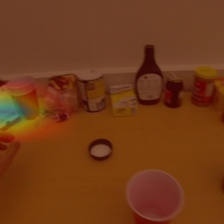}
		\includegraphics[scale=0.4]{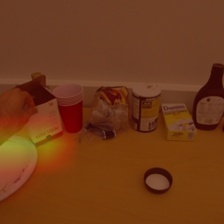}
		\includegraphics[scale=0.4]{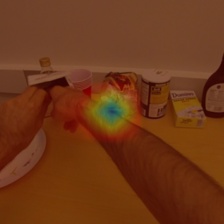}
		\includegraphics[scale=0.4]{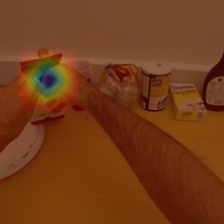}
		\includegraphics[scale=0.4]{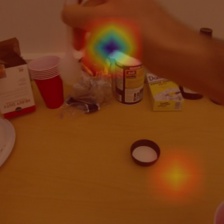}
	\end{subfigure}\\ \vskip 2mm
	\ \raisebox{.5in}{\rotatebox[origin=t]{90}{LSTA}}
	\begin{subfigure}[b]{0.95\textwidth}
		\includegraphics[scale=0.4]{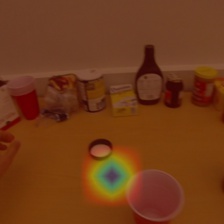}
		\includegraphics[scale=0.4]{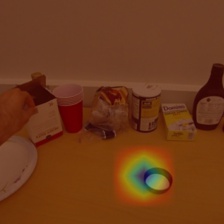}
		\includegraphics[scale=0.4]{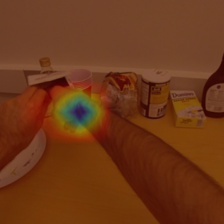}
		\includegraphics[scale=0.4]{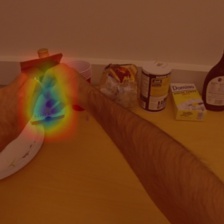}
		\includegraphics[scale=0.4]{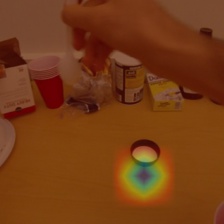}
	\end{subfigure}\\ \vskip 2mm	       
	\ \raisebox{.5in}{\rotatebox[origin=t]{90}{Flow}}
	\begin{subfigure}[b]{0.95\textwidth}
		\includegraphics[scale=0.4]{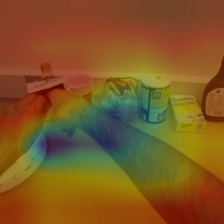}
		\includegraphics[scale=0.4]{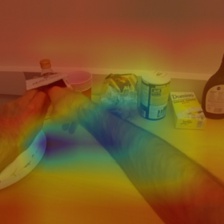}
		\includegraphics[scale=0.4]{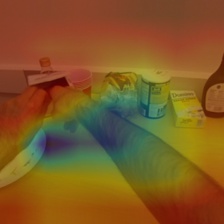}
		\includegraphics[scale=0.4]{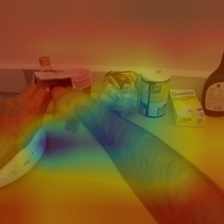}
		\includegraphics[scale=0.4]{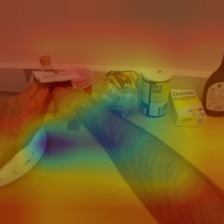}
	\end{subfigure}\\ \vskip 2mm			 
	\caption{Attention maps generated by ego-rnn (second row) and LSTA (third) for take\_spoon video sequence. We show the 5 frames that are uniformly sampled from the 25 frames used as input to the corresponding networks. Fourth row shows the attention map generated by the motion stream. For flow, we visualize the attention map on the five frames corresponding to the optical flow stack given as input.}
	\label{fig:att_map_take_spoon}
\end{figure*}

Figs~\ref{fig:att_map_scoop_sugar}~-~\ref{fig:att_map_take_spoon} visualize the generated attention maps for different video sequences. In Figs.~\ref{fig:att_map_scoop_sugar}~-~\ref{fig:att_map_shake_tea}, one can see that \acs{lsta} is able to successfully identify the relevant regions and track them across the sequences while ego-rnn misses the regions in some frames. This shows the ability of \acs{lsta} in identifying and tracking the discriminant regions that are relevant for classifying the activity category. However, in Figs.~\ref{fig:att_map_take_bread} and \ref{fig:att_map_take_spoon}, the network fails to recognize the relevant regions. In both of these video sequences, the object is not present in the first few frames and the network attends to wrong regions, failing to move its attention towards the object when it appears. Since the proposed method maintains a memory of attention maps, occlusion of the relevant object in the initial frames results in the network attending to the wrong regions in the frame.
	
\end{document}